\documentclass[runningheads]{llncs}

\usepackage{amsmath,amsfonts,bm}

\def\eqref#1{equation~\ref{#1}}

\def\1{\bm{1}}

\DeclareMathAlphabet{\mathsfit}{\encodingdefault}{\sfdefault}{m}{sl}
\SetMathAlphabet{\mathsfit}{bold}{\encodingdefault}{\sfdefault}{bx}{n}

\usepackage{hyperref}
\usepackage{url}
\usepackage{graphicx}
\usepackage{xcolor}
\usepackage{algorithm}
\usepackage{algpseudocode}
\usepackage{subfig}

\usepackage{amssymb}
\usepackage{multirow}
\usepackage{multicol}
\usepackage{ntheorem,lipsum}
\theorembodyfont{\upshape}

\usepackage{booktabs}
\usepackage{pifont}            
\usepackage[utf8]{inputenc}
\usepackage{etoolbox}
\usepackage{lscape}

\title{Dynamic Graph Representation Learning with Neural Networks: A Survey}

\author{Leshanshui YANG$^{\star,\dagger}$, Sébastien ADAM$^{\star,\ddagger}$, Clément CHATELAIN$^{\star, \mathsection}$}
\institute{$^\star$LITIS, $^\dagger$Saagie, $^\ddagger$University of Rouen Normandy, $^\mathsection$INSA Rouen Normandy} 
\date{March 2023}

\def\algbackskip{\hskip-\ALG@thistlm}

\begin{document}

\maketitle

\begin{abstract}
In recent years, Dynamic Graph (DG) representations have been increasingly used for modeling dynamic systems due to their ability to integrate both topological and temporal information in a compact representation. Dynamic graphs allow to efficiently handle applications such as social network prediction, recommender systems, traffic forecasting or electroencephalography analysis, that can not be adressed using standard numeric representations. As a direct consequence of the emergence of dynamic graph representations, dynamic graph learning has emerged as a new machine learning problem, combining challenges from both sequential/temporal data processing and static graph learning. In this research area, Dynamic Graph Neural Network (DGNN) has became the state of the art approach and plethora of models have been proposed in the very recent years.  This paper aims at providing a review of problems and models related to dynamic graph learning. The various dynamic graph supervised learning settings are analysed and discussed. We identify the similarities and differences between existing models with respect to the way time information is modeled. Finally, general guidelines for a DGNN designer when faced with a dynamic graph learning problem are provided.


\end{abstract}



\section{Introduction} \label{sec:Introduction}

Graphs are data structures used for representing both attributed entities (the vertices of the graph) and relational information between them (the edges of the graph) in a single and compact formalism. They are powerful and versatile, capable of modelling irregular structures such as skeletons, molecules, transport systems, knowledge graphs or social networks, across different application domains such as chemistry, biology or finance. This expressive power of graphs explains why graphs have been used extensively to tackle Pattern Recognition (PR) tasks, as demonstrated by the current special issue.

In the PR community, most of the existing contributions focus on static graphs where the node set, the edge set and the nodes/edge attributes do not evolve with time. Yet, for some real-world applications such as traffic flow forecasting, rumour detection or link prediction in a recommender system, graphs are asked to handle time-varying topology and/or attributes, in order to  model dynamic systems. Several terms are used in the literature to refer to graphs in which the structure and the attributes of nodes/edges evolve. Dynamic graphs \cite{rossi2020temporal,pareja2020evolvegcn}, temporal graphs \cite{singer2019node,rehman2020exploring,xu2020inductive}, (time-/temporally) evolving graphs \cite{bahmani2012pagerank,song2021temporally,hamilton2017inductive,li2019predicting}, time-varying graphs \cite{kalofolias2017learning,casteigts2012time}, time-dependent graphs \cite{wang2019time,huang2020reliable}, or temporal networks  \cite{holme2012temporal,holme2015modern,wang2021inductive,sato2019dyane} are examples of such terms which refer to conceptual variants describing the same principles. This multiplicity of terms can be explained by the diversity of the scientific communities interested by this kind of models but also by the youth of the field. It illustrates the need for precise definitions and clear taxonomies of problems and models, which is one of the contributions of this paper. In the following, we choose to use the more general term Dynamic Graph (DG). 

As a direct consequence of the emergence of dynamic graph representations, dynamic graph learning emerged as a new machine learning problem, combining challenges from both sequential/temporal data processing and static graph learning.

When learning on sequential data, the fundamental challenge is to capture the dependencies between the different entities of a sequence. In this domain, the original concept of recurrence, mainly instantiated by LSTM \cite{hochreiter1997long}, has been gradually replaced in recent years by fully convolutional architectures, which offer better parallelization capabilities during learning. More recently, sequence-to-sequence models based on the encoder/decoder framework have been proposed \cite{Sutskever14}, allowing to deal with desynchronised input and output signals. These models, such as the famous transformer \cite{vaswani2017attention}, rely on the intensive use of the attention mechanism \cite{bahdanau2014neural,Niu21,self18}.

When learning on static graphs, the main challenge is to overcome the permutation invariance/equivariance constraint inherent to the absence of node ordering in graph representations. To solve this problem, node-based message-passing mechanisms based on graph structure have led to the first generation of Graph Neural Networks (GNNs), called Message Passing Neural Networks (MPNNs) \cite{gilmer17:_neural_messag_passin_quant_chemis}. As convolutions on images, these models propagate the features of each node to neighbouring nodes using trainable weights that can be shared with respect to the distance between nodes (Chebnet) \cite{defferrard16:_convol_neural_networ_graph_fast}, to the connected nodes features (GAT) \cite{velivckovic2017graph} and/or to edge features (k-GNN) \cite{bresson2017residual}. Given the maturity of such models and their applicability for large sparse graphs, they have been applied with success on many downstream tasks. This maturity also explains the existence of some exhaustive reviews and comparative studies, such as in  \cite{zhou2020graph,wu2020comprehensive,xu2018how,dwivedi2020benchmarking} to cite a few. One can note that, despite these successes, it has been shown that MPNNs are not powerful enough \cite{xu2018how}. That is why machine learning on graphs is still a very active field, trying to improve the expressive power of GNNs \cite{morris2019wl,maron2019provably,breakingthelimits}. However, these models are still too computationally demanding to be applicable to real-world problems. 

Compared to neural networks applied for learning on sequences and on static graphs, Dynamic Graph Neural Network (DGNN) is a much more recent field. To the best of our knowledge, the founding DGNN models are from 2018 \cite{seo2018structured} and 2019 \cite{trivedi2019dyrep} for respectively the discrete and the continuous cases. These papers have been at the root of a "zoo" of methods proposed by various scientific communities, with various terminologies, various learning settings and various application domains.   

In order to structure the domain, some state-of-the-art papers have been published recently \cite{kazemi2020representation,skarding2021foundations,barros2021survey}. Without focusing on DGNN, these papers give a good overview of many machine learning issues linked to dynamic graphs and describe many existing models. However, they do not compare models according to the different possible DG inputs and supervised learning settings that can be encountered when applying machine learning to DG. As an example, \cite{kazemi2020representation,skarding2021foundations} do not consider in their study the spatial-temporal case. They also do not compare the ability of existing models to consider inductive or transductive tasks. 

The main objective of this review is to extend the existing studies mentioned above, focusing on \textbf{dynamic graph supervised learning using neural networks}. It is addressed to the audience with fundamental knowledge of neural networks and static graph learning. Three main contributions can be highlighted. The first one consists in \textbf{clarifying and categorizing the different dynamic graph learning contexts} that are encountered in the literature. These contexts are distinguished according to the type of input DGs (discrete vs. continuous, edge-evolving vs. node-evolving vs attributes-evolving, homogeneous vs. heterogeneous) but also according to the learning setting (transductive vs. inductive). The second contribution is an \textbf{exhaustive review of existing DGNN models}, including the most recent ones. For this review, we choose to categorize models into five groups, according to the strategy used to incorporate time information in the model, which is the main challenge for the application of neural networks on DGs. Based on this categorisation, and using the taxonomy of contexts mentioned above, the third contribution is to provide some \textbf{general guidelines for a DGNN designer when faced with a dynamic graph learning problem} and to describe different methods for optimising the DGNN performance.  

The remainder of this paper is structured as follows. Section 2 relates to the first contribution, by considering the inputs, the outputs and the learning settings that can be encountered when learning on dynamic graphs. Section 3 reviews existing Dynamic Graph Neural Networks (DGNNs) and compares them according to their temporal information processing. Finally, section 4 brings forward the guidelines for designing DGNNs and discusses some optimisation methods.

\section{Dynamic Graph Representation Learning}
\label{sec:2}

\begin{figure}[!h]
	\centering
	\includegraphics[width=0.7\columnwidth]{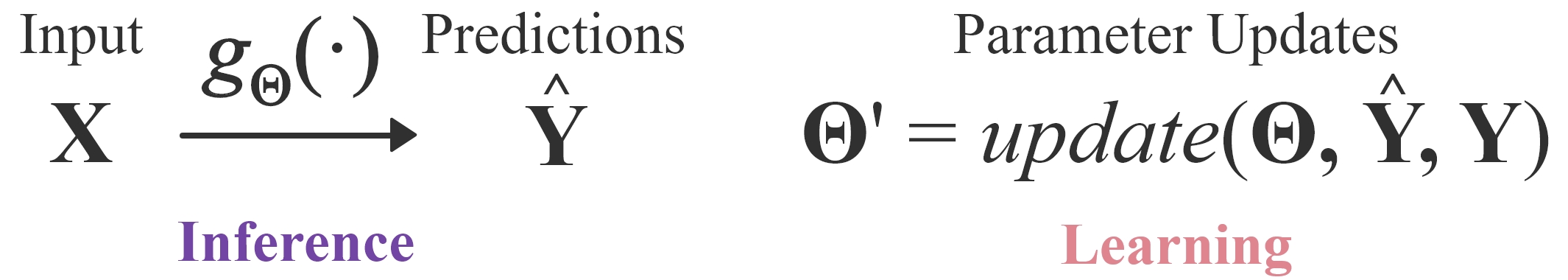}
	\caption{Left: inference phase for making predictions $\hat{\textbf{Y}}$ on given data \textbf{X}. Right: learning phase for updating the parameters $\Theta$ of the predictor $g$.}
	\label{fig:2suplearn}
\end{figure}

Regardless of the data representation, the goal of supervised learning methods is to build a parameterized statistical model or predictor $g_{\Theta}$ that maps between an input space $\mathcal{X}$ and an output space $\mathcal{Y}$ (see Fig. \ref{fig:2suplearn}). During the \textit{learning} phase, the training of the predictor $g_{\Theta}$ consists in updating its parameters $\Theta$ using a dataset of couples $(x,y)$. The update is computed by a minimization of the loss between the prediction $\hat{\textbf{Y}}$ and the ground truth $\textbf{Y}$.

\begin{figure}[h!]
    \centering
    \includegraphics[width=0.8\columnwidth]{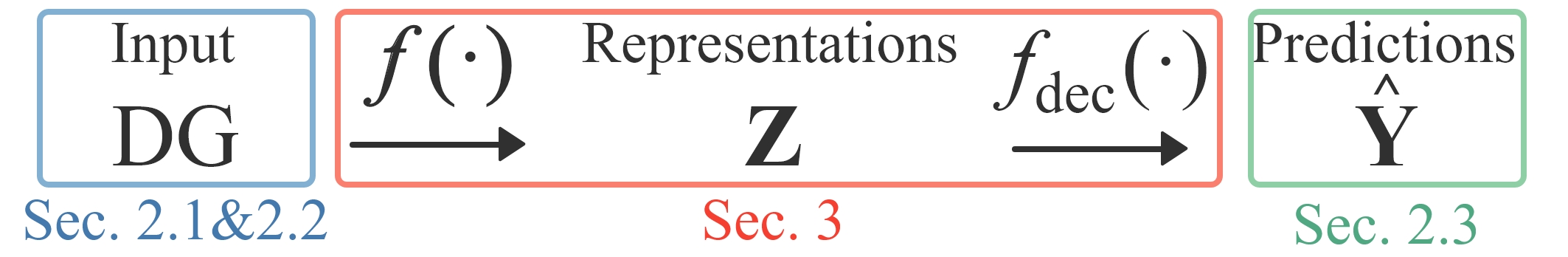}
    \caption{Encoder/decoder model applied on dynamic graphs: the encoding consists in computing $\textbf{Z} = f(\mathrm{DG})$, where DG is a dynamic graph (including both topology and attributes), $f(\cdot)$ is a parameterized statistical model (typically a neural networks with learnable parameters), and \textbf{Z} is the encoded tensor representation of DG. The decoder $f_{dec}(\cdot)$ takes as input the representations $\textbf{Z}$ to get the predictions $\hat{\textbf{Y}}$. }
    \label{fig:2article_struc}
\end{figure}

Many recent models follow the encoder/decoder principle, where a variable-length input signal is encoded into a latent representation, which is then used by a decoder to compute the output signal for the downstream task (see Fig. \ref{fig:2article_struc}). The fixed-size latent representation allows alignment between variable-size input and output signals that are not necessarily synchronised, i.e. the units of the input and output sequences may have a different order.
It is of great interest in many sequence-to-sequence problems involving text, images or speech. 
The learning latent representation (also known as \emph{embedding}) is called \textit{representation learning}.



Sequences are signals in which information varies according to one or more dimensions, that generally define a position in a structure. This structure may have one or multiple dimensions: time for speech, 1D position for text, 2D position for images, etc.
In the case of dynamic graphics, the information varies according to the position too, but the structure can also evolve along time. In this section, we propose the concept of \emph{degree of dynamism} that defines the nature of the variability. Given that the degree of dynamism of DG is higher than for sequences, the encoder/decoder framework therefore provides a very suitable framework for learning DG representations.

In this section, we define important concepts specific to dynamic graphs, giving the necessary material for understanding the review of the dynamic graph embedding problem presented in section \ref{sec:DynamicGraphNeuralNetworks}. The section
is structured as follows (see Fig. \ref{fig:2article_struc}): after giving useful definition about static graphs in subsection \ref{sec:2_staticgrepr}, we present the representation of dynamic graphs in subsection \ref{sec:2_dgrepr}. We then present the output shape and the transductive/inductive nature of learning tasks in subsections \ref{sec:2_output} and \ref{sec:2_cases}. Finally, we introduce the related applications in each learning setting in subsection \ref{sec:2_tasks}.

\subsection{Static Graph Modeling} \label{sec:2_staticgrepr}
A static graph $G$ can be represented topologically by a tuple $(V, E)$ where $V$ is the node set of $G$ and $E$ is the edge set of $G$. The connectivity information is usually represented by an adjacency matrix $\textbf{A} \in \mathbb{R}^{|V| \times |V|}$. In this matrix, $A(u,v)=1$ if there is an edge between node $u$ and $v$, $A(u,v)=0$ otherwise. \textbf{A} is symmetric in an undirected graph, while in the case of directed graphs \textbf{A} is not necessarily symmetric. 

Nodes usually have attributes represented by a feature matrix $\textbf{X}_{V} \in \mathbb{R}^{|V| \times d_{V}}$, where $d_V$ is the length of the attribute vector of a single node. Similarly, edges may have attributes (such as weights, directions, etc.) which can also be represented by a matrix $\textbf{X}_{E} \in \mathbb{R}^{|E| \times d_{E}}$.

In the case of weighted graphs, the values in the matrix \textbf{A} are the weights of each edge instead of $1$ which is denoted as:

$$ \textbf{A}_{u,v}=\left\{
\begin{array}{rcl}
w_{u,v}       &      & {\textrm{if }   (u,v) \in E}\\
0     &      & \textrm{otherwise} .
\end{array} \right. $$

For some specific applications, both nodes and edges can be of different types. For example, in recommender systems, nodes can usually be mapped into two types, item and user, and have feature matrices of different sizes and contents. In such cases, we extend the notation of the graph $G= (V, E)$ with the type mapping functions $\phi: V \rightarrow O$ and $\psi:E \rightarrow R$, where $|O|$ denotes the possible types of nodes and $|R|$ denotes the possible types of edges \cite{wang2019heterogeneous}. When $|O|=1$ and $|R|=1$, nodes and edges are of a single type, this graph is called homogeneous. In contrast, in a heterogeneous graph, $|O|+|R|>2$ and each type of node or edge could have its own number of feature dimensions \cite{wang2019heterogeneous,luo2020dynamic}. 

\begin{figure*}[h]
    \centering
    \includegraphics[width=\textwidth]{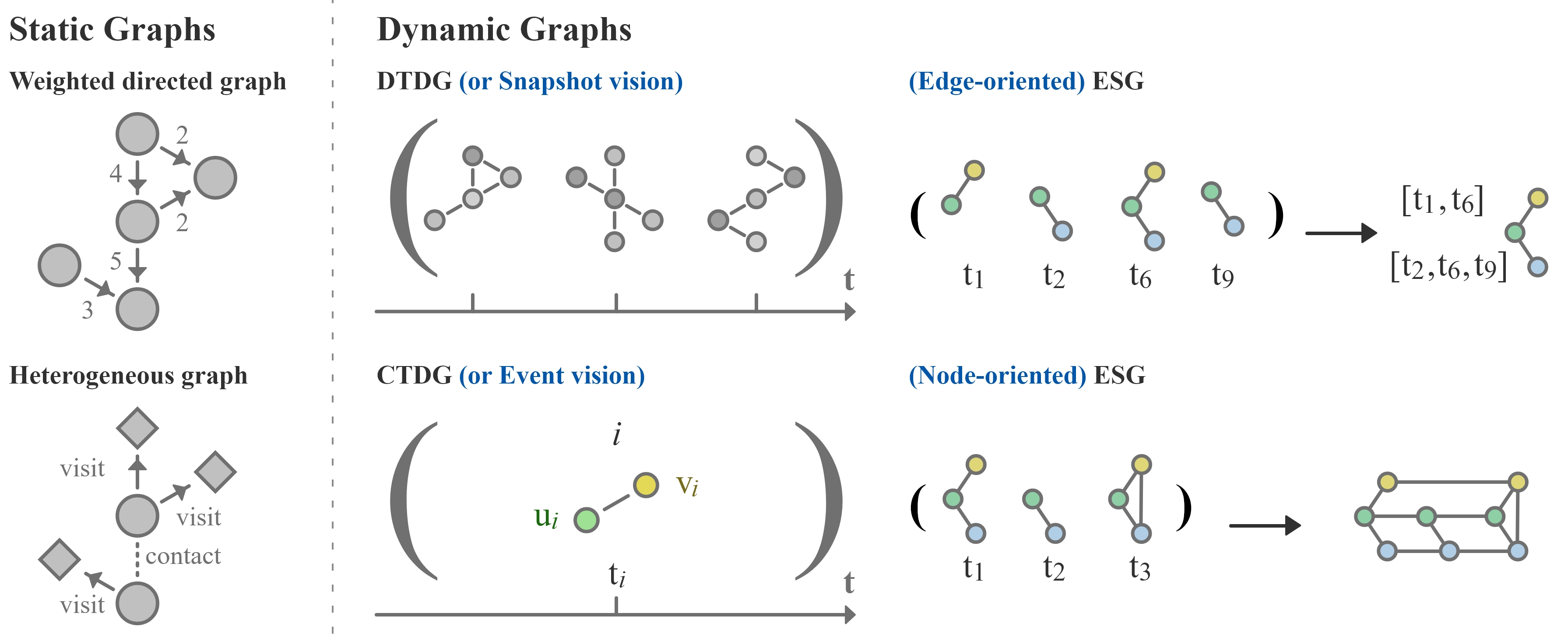}
    \caption{Left: Common static graph representations. The edges in a directed weighted graph have directions and weights, which are frequently used when modelling email networks, citation networks, etc. The nodes and edges in a heterogeneous graph can have multiple possible types i.e., recommendation systems and knowledge graphs. Middle: Discrete Time Dynamic Graph (DTDG) represented by snapshots, and Continuous Time Dynamic Graph (CTDG) represented by events, the example in the figure is the "Contact Sequence" case. Right: Equivalent Static Graphs (ESGs) represented by edges and nodes.}
    \label{fig:2_g_repr}
\end{figure*}

Figure \ref{fig:2_g_repr} illustrated the difference between homogeneous and heterogeneous graphs.

\subsection{Dynamic Graph Modeling} \label{sec:2_dgrepr}

~\\
\begin{definition}[Dynamic Graph]
A dynamic graph is a graph whose topology and/or attributes change over time.
\end{definition}

According to this definition, both structure and attributes may change over time in a dynamic graph. Edges and/or nodes may be added or deleted and their attributes may change. Thus, this definition covers different configurations. In order to distinguish between them, we propose the concept of \emph{degree of dynamism} defined as follows:  

\begin{definition}[Degree of Dynamism (Node-Cintroduceentric)] \label{def:2degdyn}
The \emph{degree of dynamism} of a DG describes whether the topology, i.e. the edge set $E$ and the node set $V$, are invariant. Theoretically, there are 4 possible situations: (1) Both $V$ and $E$ are invariant, denoted as $fix_{V, E}$. This situation corresponds to DG called  "Spatial-Temporal Graphs" or "Spatio-Temporal Graphs (STGs) in the literature
(2) $V$ is invariant but the edge set is changing, denoted as $fix_{V}$. (3) The node set and the edge set are both changing, denoted as $vary$. (4) The set of edges is constant but the set of nodes changes. Since an edge exists based on a tuple of nodes, this situation is meaningless.
\end{definition}

The Table \ref{tab:2degdyn} illustrates these different configurations which will be discussed throughout the paper. 

\begin{table}[htbp!]\centering
\caption{Degree of Dynamism discussed in the article.}\label{tab:2degdyn}

\begin{tabular}{c c c c}\toprule
\multicolumn{2}{c}{\multirow{2}{*}{}} &\multicolumn{2}{c}{Node Set} \\\cmidrule{3-4}
& &Constant &Variant \\\cmidrule{1-4}
\multirow{2}{*}{Edge Set } &Constant &\textit{$fix_{V, E}$} & N/A \\\cmidrule{2-4}
&Variant &\textit{$fix_{V}$}  &\textit{$vary$}  \\\midrule
\bottomrule
\end{tabular}
\end{table}

 In the following, we provide the main existing dynamic graph representations for DG, synthesizing previous studies that focus on subparts of these representations \cite{wang2019time,holme2012temporal,zaki2016comprehensive,skarding2021foundations,wu2019graph,gao2022equivalence}. We also introduce in this part a representation called Equivalent-Static-Graph which consists in modeling a DG using a static graph. The different models are compared in Fig. \ref{fig:2_g_repr}.

\subsubsection{Continuous Time Dynamic Graphs}
To preserve accurate time information, Continuous Time Dynamic Graphs (CTDGs) use a set of events to represent dynamic graphs. 
Existing work \cite{skarding2021foundations} mainly focuses on the dynamics of edges and outlines three typical representation methods that represent events with index $i=1,2,\ldots$ by giving a pair of nodes $(u_{i}, v_{i}$) and time $t_{i}$: 

\textit{Contact-Sequence} \cite{holme2012temporal,skarding2021foundations} is used to represent the instantaneous interaction between two nodes $(u,v)$ at time $t$.

\begin{equation}
    \textit{Contact-Sequence} = \left\{(u_i, v_i, t_i)\right\}
\end{equation}

\textit{Event-Based} \cite{skarding2021foundations,guo2022continuous} dynamic graphs represent edges with a time $t_i$ and duration $\Delta_i$. They are similar to the \textit{Interval-Graph} defined in \cite{holme2012temporal}. The difference is that \textit{Interval-Graph} uses a set $T_e$ of start and end times $(t_i, t_i')$ to represent all active times of the edge, rather than the duration of each interaction in \textit{Event-Based}.

\begin{equation}
    \textit{Event-Based} = \left\{(u_i, v_i, t_i, \Delta_i)\right\}
\end{equation}
\begin{eqnarray}
    \textit{Interval-Graph} = \left\{(u_i, v_i, T_e); T_e=\left((t_1,t_1'), (t_2,t_2'), ...\right)\right\}
\end{eqnarray}

\textit{Graph Stream} \cite{skarding2021foundations} is often used on massive graphs \cite{mcgregor2014graph,zhang2010survey}. It focuses on edges' addition ($\delta_i=1$) or deletion ($\delta_i=-1$).  

\begin{equation}
    \textit{Graph Stream} = \left\{(u_i, v_i, t_i, \delta_i)\right\}
\end{equation}

\subsubsection{Discrete Time Dynamic Graphs}
Discrete-time dynamic graphs (DTDGs) can be viewed as a sequence of $T$ static graphs as shown in Eqn. \ref{2-eq-dt}. They are snapshots of the dynamic graph at different moments or time-windows. DTDGs can be obtained by periodically taking snapshots of CTDGs on the time axis \cite{zaki2016comprehensive,skarding2021foundations}.
\begin{equation} \label{2-eq-dt}
    \textit{DTDG} = (G^1, G^2, ..., G^T)
\end{equation}


\subsubsection{Equivalent Static Graphs} \label{sec:2_2_ESG}
Representations of this category consist in constructing a single static graph, called Equivalent Static Graph (\textit{ESG}), for representing a dynamic graph. Several methods for constructing \textit{ESG} have been proposed in recent years. We divide them into two categories: \textit{edge-oriented} and \textit{node-oriented ESG}.

\textit{Edge-oriented ESG} aggregates graph sequences into a static graph with time information encoded as sequences of attributes \cite{rehman2020exploring,holme2012temporal,holme2015modern} as shown in Fig. \ref{fig:2_g_repr} (right top). Such representation is also called \textit{time-then-graph} representation \cite{gao2022equivalence}.

\textit{Node-oriented ESG} builds copies of vertices at each moment of their occurrence and defines how the nodes are connected between timestamps/occurrences \cite{rehman2020exploring,azimifar2015vehicle,li2022enhanced,DBLP:journals/corr/Michail15}. A simple example is shown in Fig. \ref{fig:2_g_repr} (right bottom), further details are discussed in subsection 3.1. 

A major interest of ESG representations is that they make static graph algorithms available for learning on dynamic graph.


\subsection{Dynamic Graph Output Granularity} \label{sec:2_output}
As said before, supervised learning aims to learn a mapping function between an input space $\mathcal{X}$ and an output space $\mathcal{Y}$. 
The previous subsection introduces the space in which a DG $x \in \mathcal{X}$ can be represented. This subsection now discusses the space $\mathcal{Y}$. 

Since a dynamic graph involves concepts from both graph and temporal/sequential data, both aspects must be considered to define the space $\mathcal{Y}$. 

When learning a statistical model on sequential data, the input \textbf{X} can be represented by the $d$-dimensional features at $T$ time steps, denoted as $\textbf{X} \in \mathbb{R}^{T \times d}$.  
The granularity of the outputs \textbf{Y} can be either timestep-level (one output per time step, as in \textit{Part of speech} tagging or more generally sequence labelling), or aggregated (one output for many inputs, as in sentiment classification or more generally sequence classification).

In static graph learning, the inputs have topological information in addition to the features  $\textbf{X}_V$ and $\textbf{X}_E$. As for sequences, the outputs \textbf{Y} can be local (one label per node or per edge) or global (one label per subgraphs or for the entire graph).

As a consequence, the output granularity when considering DG can be temporally timestep-level or aggregated, and topologically  local or global. 

\subsection{Transductive/Inductive on Dynamic Graphs} \label{sec:2_cases}
When learning on static graphs, transductive and inductive tasks are frequently distinguished. Transductive tasks consist in taking decision for a set $V_{inference}$ of unlabeled nodes, the model being learnt on a set of labeled nodes $V_{learning}$ of the same graph. In such a situation, the features and the neighborhood of nodes in $V_{inference}$ can be exploited by the learning algorithm in an unsupervised way. That is why this situation is also called a semi-supervised learning on graphs \cite{van2020survey}. In contrast, an inductive task is the case when there exist nodes in the $V_{inference}$ that have not been seen during the learning phase \cite{hamilton2017inductive}. This situation typically occurs when learning and inference are performed on different static graphs.

In the case of dynamic graphs, the evolving nature of both $V$ and $E$ brings more possible scenarios when considering the transductive/inductive nature of the tasks. Various configurations should be defined, considering the "degree of dynamism" defined in section \ref{def:2degdyn}. We use the term "case" to distinguish between different transductive/inductive natures of dynamic graph learning taking into account the degree of dynamism.

Specifically, we divide learning on dynamic graphs into five cases: 
\begin{itemize}
    \item \textit{Trans-fix$_{V, E}$} (1) : in this case, the topology of the DG is fixed, $V$ and $E$ are therefore the same on the learning and inference sets. 
    \item \textit{Trans-fix$_{V}$} (2) : in this case, the learning and inference node sets are fixed and equal but the edge set evolves, which requires taking into account the evolving connectivity between nodes on the graph.
    \item \textit{Trans-vary} (3) : in this case, $V_{learning}^t$ and $V_{inference}^t$ may evolve over time, but the presence of each node to be predicted in the test set are already determined in the learning phase : $\forall v \in V_{inference}, v \in DG_{learning}$. 
    \item \textit{Ind$_{V}$} (4) : this case refers to node-level inductive tasks, where the learning and inference are on the same DG, but $\exists v \in V_{inference}, v \notin DG_{learning}$. Although the label and historical attributes from the learning phase can be reused on the inference set, a major challenge is to handle the unseen nodes and the uncertain number of nodes. 
    \item \textit{Ind$_{DG}$} (5) : this case refers to DG-level inductive learning, where learning and inference are performed on different DGs. The statistical model needs to handle the complete unseen dynamic graphs. 
\end{itemize} 

\begin{figure}[!h] \label{fig:2_cases}
	\centering
	\includegraphics[width=.8\columnwidth]{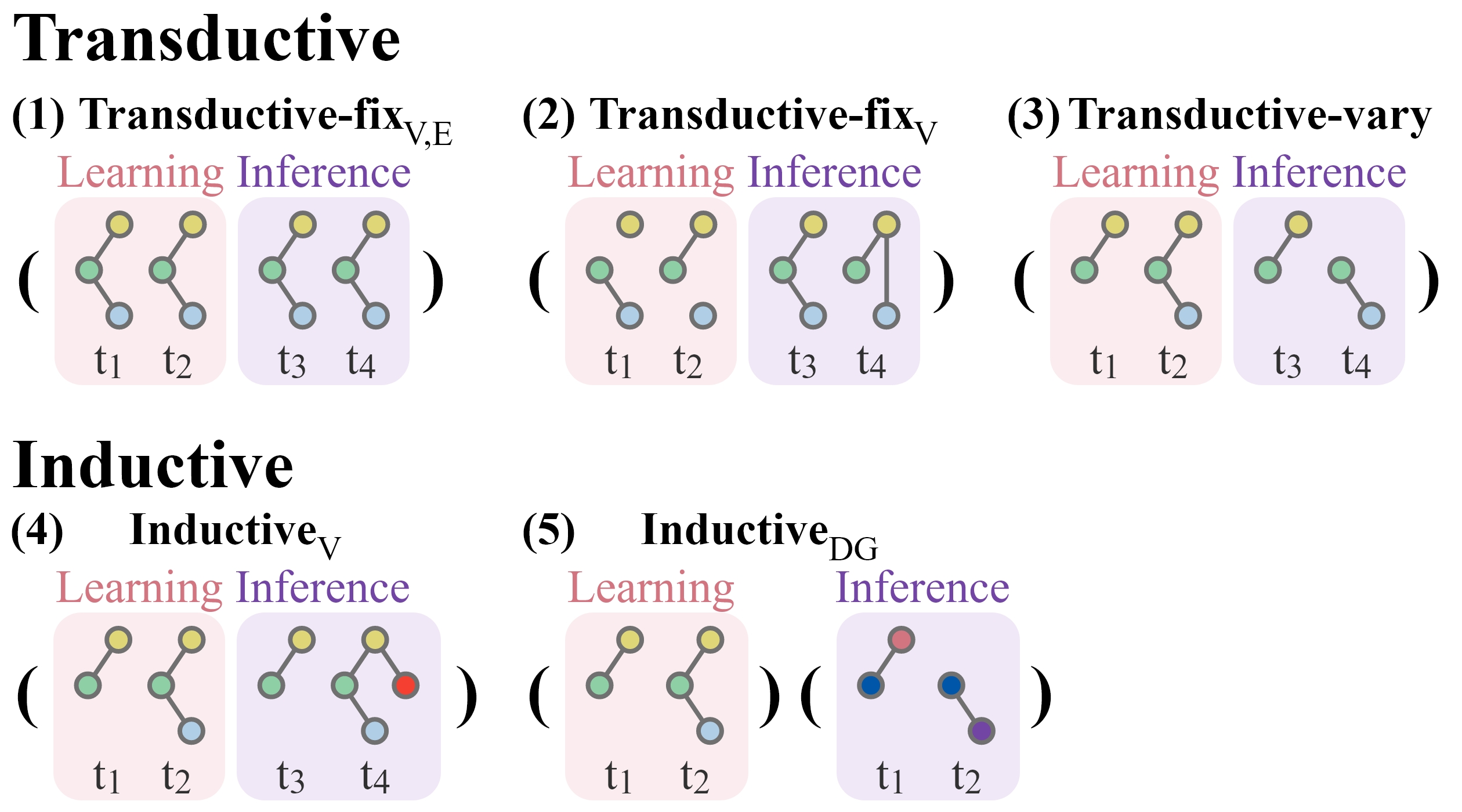}
	\caption{The transductive and inductive cases in dynamic graph learning under discrete time: (1) denotes the case where both the node set and edge set are fixed in learning and inference. (2) denotes the case where only the node set is fixed for learning and inference. (3) denotes the case where the node set changes but no unseen node appears for inference. (4) denotes the inductive case where the train and test are on the same DG but there are new nodes for inference. (5) denotes the inductive case where there are new DGs for inference.}
	\label{fig:2case}
\end{figure}

These 5 cases are illustrated in Fig. \ref{fig:2case}. 



\subsection{Dynamic Graph Predictive Tasks} \label{sec:2_tasks}
In the previous subsections, we have investigated dynamic graph learning specificities related to the representations of the input (discrete vs. continous time), the granularity of the output (local vs. global, timestep-based vs. aggregated), and different learning \textit{cases}. In this section, we categorize existing contributions in the literature according to these four criteria.

Table \ref{tab:task} gives a synthetic view of this categorization with an emphasis on the targeted applications and on the metrics used for assessing models performance.

As one can see in this table, in \textbf{Discrete Time Transductive Tasks}, related applications usually concern relatively stationary topologies, i.e. \textit{Trans-fix$_{V,E}$}, such as human body structure or geographic connectivity. Some typical node-level/local tasks predict attributes for the next time step(s) based on past time step(s), such as traffic flow \cite{wu2019graph,yu2017spatio,li2017diffusion,ruiz2020gated,wu2020connecting}, number of infectious disease cases \cite{deng2020cola,kapoor2020examining,wang2022causalgnn}, number of crimes \cite{xia2022spatial} and crop yields \cite{fan2022gnn}. Graph-level/global tasks either retrieve the class of each snapshot like sleep stage classification \cite{jia2020graphsleepnet} or output a prediction for the entire DG such as the emotion of a skeletal STG \cite{chaolong2018spatio}. When the DG has a fixed set of nodes with evolving edges, i.e. \textit{Trans-fix$_{V}$}, such \textit{cases} can be employed for modelling the connectivity in the telecommunication network \cite{li2019predicting} or contact of individuals in a conference \cite{sato2019dyane}.

In \textbf{Discrete Time Inductive Tasks}, i.e. when unseen nodes need to be predicted in discrete time i.e. \textit{Ind$_{V}$}, some typical tasks concerns node classification \cite{singer2019node} or link prediction \cite{singer2019node,sato2019dyane,zhou2019dynamic} for future snapshots in social networks. An example of \textit{Ind$_{DG}$} with graph-level output is classifying real and fake news based on the snapshots of its propagation tree on the social network \cite{choi2021dynamic,sun2022ddgcn,wang2020fake}.

Dynamic graphs in \textbf{Continuous Time} are widely used to model massive dynamic graphs that frequently have new events, such as recommendation systems \cite{tang2021dynamic,li2020dynamic,kumar2019predicting} or social networks in the transductive \textit{cases} \cite{qu2020continuous,nguyen2018continuous} or inductive \textit{cases} \cite{rossi2020temporal,singer2019node,xu2020inductive,wang2021inductive,trivedi2019dyrep}. Local tasks predict the properties and interactions of seen or unseen nodes, under transductive and inductive \textit{cases}, respectively. Since CTDGs have no access to global/entire graph information under their minimum time unit, the global timestep level label is meaningless under continuous time. However, global aggregated tasks can be implemented by aggregating nodes of different time steps, such as rumour detection in continuous time \cite{song2021temporally}.  
Since CTDG can be transformed to DTDG by periodically taking snapshots \cite{zaki2016comprehensive,skarding2021foundations}, the above tasks can also be considered as tasks under dynamic time with a lower time resolution.

To evaluate the performance of a statistical model on a given task, traditional machine learning metrics are employed as shown in table \ref{tab:task}. When the output predictions are discrete values, i.e. for \textbf{classification} task \cite{rossi2020temporal,singer2019node,xu2020inductive,song2021temporally,trivedi2019dyrep}, common metrics include accuracy, precision, recall, F1, and area under the receiver operating characteristic (AUROC). When the output values are continuous values, i.e. for \textbf{regression} task \cite{wu2019graph,yu2017spatio,li2017diffusion}, common metrics are mean absolute (percentage) error, root mean square (log) error, correlation, and R squared. \textbf{Node ranking} tasks \cite{li2020dynamic,kumar2019predicting} predict a score for each node and then sort them. These tasks can be evaluated by the reciprocal rank, recall@N, cumulative gain and their variants.
Note that dynamic tasks are generally evaluated using static metrics computed along the time axis.
\begin{landscape}

\begin{table*}[h]\centering
\caption{Common dynamic graph predictive tasks and related applications categorized by their input time granularity, transductive/inductive cases, and label shapes. "Clf." indicates classification and "Reg." indicates regression. Social networks refers to contact network and citation network, whether it is heterogeneous or not. Where "DT" refers to "Discrete time" and "CT" refers to "Continuous time". 
 }\label{tab:task}






\scriptsize
\begin{tabular}{lllllllll}\toprule
Input & Case & Label & Label time & Application & Clf. & Reg. & Metrics & Publications \\
time & & object & & & & & & \\\midrule

DT & \textit{Trans-fix$_{V,E}$} & Global & Aggregated & Video Clf. & \checkmark & & Acc., AUROC& \cite{duta2020dynamic,nicolicioiu2019recurrent}\\
 & \textit{Trans-fix$_{V,E}$} & Global & Aggregated & EEG-based task and gender clf. & \checkmark & & Acc., AUROC& \cite{kim2021learning}\\
&&&& Skeleton-based emotion clf. &\checkmark & & Acc. &\cite{chaolong2018spatio}\\
 & \textit{Trans-fix$_{V,E}$} & Global & Timestep & EEG-based sleep stage clf. &\checkmark & & Acc., F1 & \cite{jia2020graphsleepnet}\\
 & \textit{Trans-fix$_{V,E}$} & Local & Both & Crop yield prediction & & \checkmark& RMSE, Corr., R² & \cite{fan2022gnn}\\
&&&& Traffic flow forecasting& & \checkmark & MAE, MAPE, RMSE & \cite{wu2019graph,yu2017spatio,li2017diffusion,ruiz2020gated,wu2020connecting}\\
&&&& Influenza forecasting  & & \checkmark & RMSE, MAE, Corr & \cite{deng2020cola,kapoor2020examining,wang2022causalgnn}\\
&&&& Crime prediction  & \checkmark & \checkmark & F1 & \cite{xia2022spatial}\\
 & \textit{Trans-fix$_{V}$} & Local & Both & Social net. link prediction &\checkmark& & AUC, Pr., Rec., F1  & \cite{singer2019node,zhou2019dynamic,chen2018gc,chen2019lstm} \\
 & \textit{Trans-fix$_{V}$} & Local & Both & Social net. Node clf. &\checkmark& &  AUC, F1 & \cite{singer2019node,sato2019dyane} \\
 & \textit{Trans-fix$_{V}$} & Local & Both & Path (edge set) availability clf. &\checkmark& & Pr., Rec., F1 & \cite{li2019predicting} \\
 & \textit{Trans-vary} & Local & Aggregated & Social net. node clf. &\checkmark& & Acc., F1  & \cite{manessi2020dynamic,wang2020generic} \\
 & \textit{Trans-vary} & Local & Aggregated & Anomalous edges clf. &\checkmark& & AUC & \cite{behrouz2022anomaly} \\

\hline

DT & \textit{Ind$_{V}$} & Local & Both & Social net. link prediction  &\checkmark& & AUC, Pr., Rec., F1 & \cite{pareja2020evolvegcn,hajiramezanali2019variational}\\
 & \textit{Ind$_{DG}$} & Global & Aggregated & Rumor detection &\checkmark& & Acc., Pr., Rec., F1 & \cite{choi2021dynamic,sun2022ddgcn,wang2020fake}\\\hline

CT & \textit{Trans-vary} & Local & Timestep & Social net. link prediction &\checkmark& & AP, Acc. & \cite{qu2020continuous,nguyen2018continuous} \\\hline

CT & \textit{Ind$_{DG}$} & Global & Aggregated & Rumor detection &\checkmark& & Acc., F1 & \cite{song2021temporally} \\
& \textit{Ind$_{V}$} & Local & Timestep & Social net. link prediction &\checkmark& & AP, Acc. & \cite{rossi2020temporal,singer2019node,xu2020inductive,wang2021inductive,trivedi2019dyrep,kumar2019predicting} \\

&&&& Social net. Node Clf. &\checkmark& & ROC AUC &\cite{xu2020inductive,rossi2020temporal,kumar2019predicting} \\
&&&& Rating prediction  & &\checkmark& MAE, RMSE & \cite{tang2021dynamic} \\
&&&& Event time prediction  & & \checkmark& MAE & \cite{trivedi2019dyrep} \\
&&&& Item recommendation & & \checkmark& Recall@10, MRR & \cite{li2020dynamic,kumar2019predicting} \\

\bottomrule
\end{tabular}
\end{table*}

\end{landscape}
\section{Dynamic Graph Embedding with Neural Networks}
\label{sec:DynamicGraphNeuralNetworks}

In the previous section, we have introduced various DG predictive tasks settings and we have categorized literature contributions according to these settings. In this section, we now take the model point of view, by describing how DGNN embed dynamic graphs into informative vectors for subsequent predictions. 

We first introduce the general idea of dynamic graph embedding in the context of the different learning tasks mentioned in subsection \ref{sec:2_tasks}. We then dive into different embedding approaches, by categorizing them according to the strategy used for handling both temporal and structural information. Finally, we present the methods for handling heterogeneous dynamic graphs.

From an encoder-decoder perspective, a deep learning statistical model first maps the original input into embeddings denoted as \textbf{Z}, and then exploits \textbf{Z} to predict an output \cite{kazemi2020representation,hamilton2017representation}. 
Graphs can be embedded either at node/edge-level or at (sub)graph-level \cite{barros2021survey,cai2018comprehensive}. Node-level embedding benefits a wide range of node-related tasks and allows more complete input information to be retained for later computation \cite{cai2018comprehensive}. In the same way, time-step level embedding retains more information than time-aggregated embedding Similarly when learning on sequential data \cite{xue2022dynamic}. 

\begin{figure}[h]
    \centering
    \includegraphics[width=.6\columnwidth]{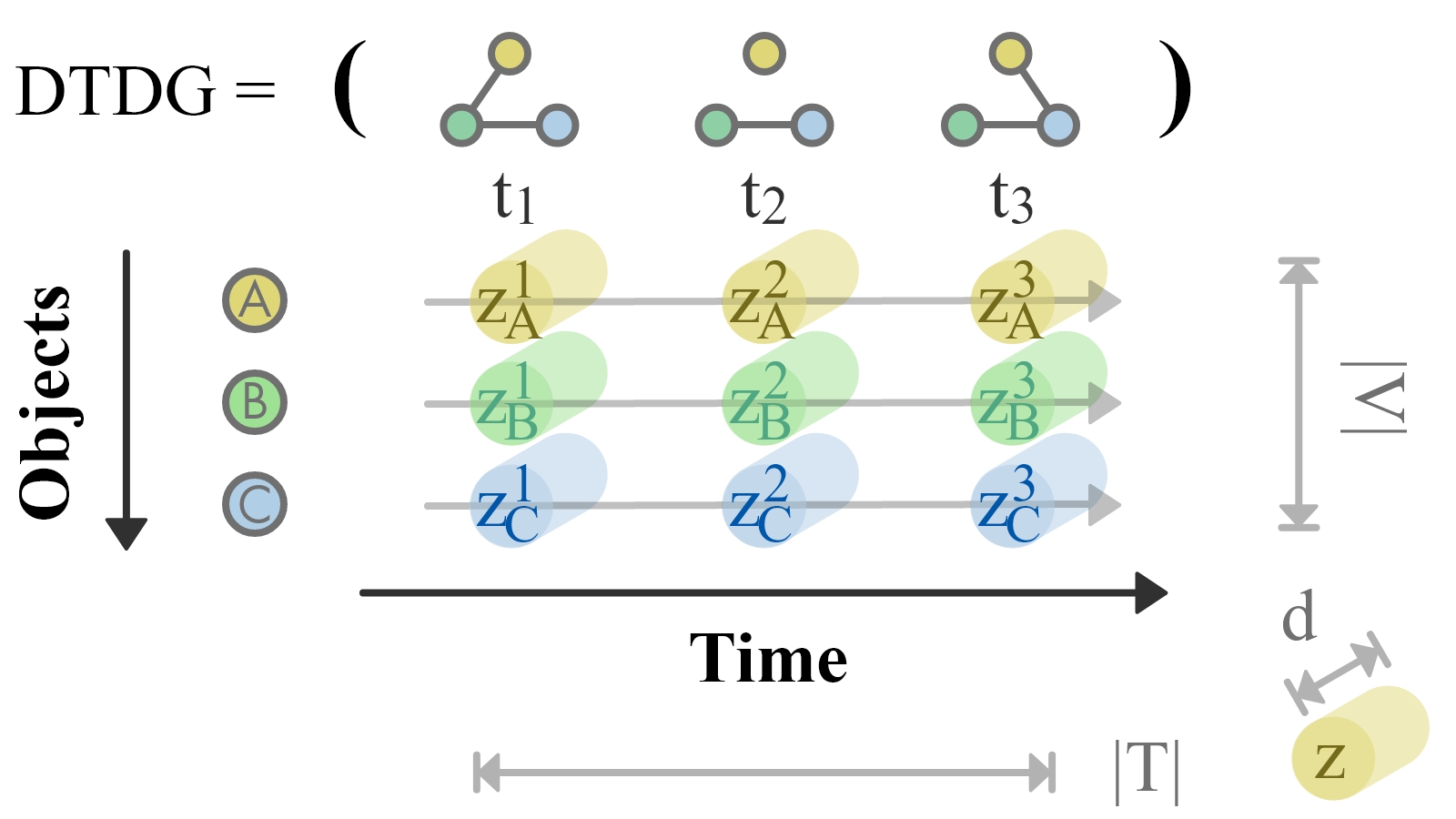}
    \caption{The most fine-grained node embedding $\textbf{Z} \in \mathbb{R}^{|V| \times |T| \times d}$, where $|V|$ is the number of nodes, $|T|$ is the number of timesteps, and $d$ is the dimension of embedding $z_{v}^{t}$ of a single node $v$ at a single timestep $t$.}
    \label{fig:2_5_emb_stg}
\end{figure}

As a consequence, embedding a dynamic graph at its finest granularity consists in computing a $d$-dimensional vector representation $\textbf{z}_v^t \in \mathbb{R}^d$ for each node $v \in V$, at all time steps $t \in T$. In this case, the embedding of the dynamic graph is given by $\textbf{Z} \in \mathbb{R}^{|V| \times |T| \times d}$, as shown in Fig. \ref{fig:2_5_emb_stg}. 

\begin{figure}[h]
    \centering
    \includegraphics[width=0.9\columnwidth]{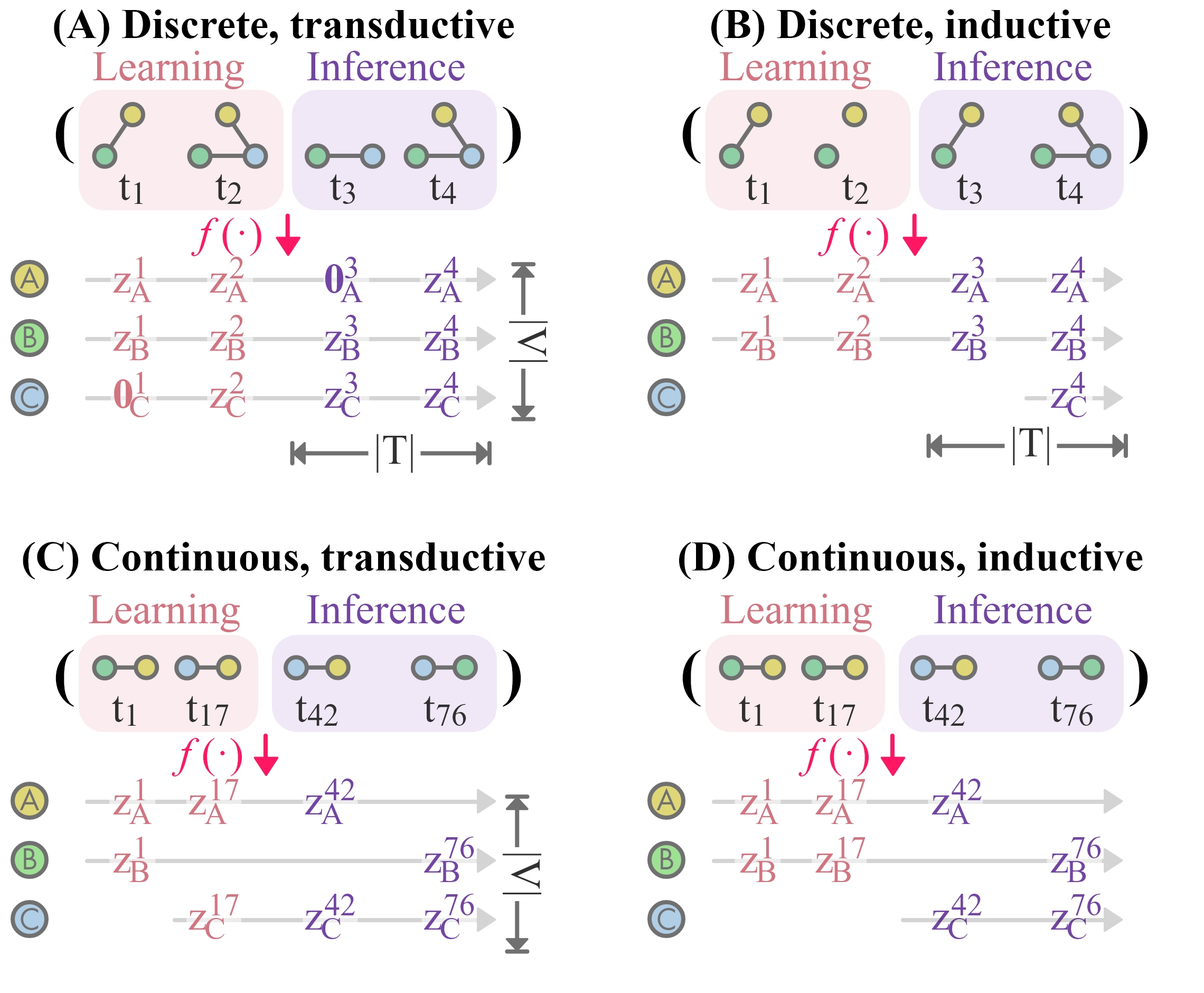}
    \caption{The available most fine-grained embedding under different dynamic graph learning settings.}
    \label{fig:2_5_emb}
\end{figure}

However, the different input time granularity and learning settings mentioned in the previous section do not always enable such an "ideal" embedding $\textbf{Z}$. In this subsection, we generalise the practicable embeddings under these different settings as shown in Fig. \ref{fig:2_5_emb} and Tab. \ref{tab:2set}. 

For discrete time transductive settings, when the node set is constant, (i.e. for \textit{cases} (1) \cite{yu2017spatio,li2017diffusion,chaolong2018spatio} and (2) \cite{singer2019node,zhou2019dynamic,chen2018gc}), the input nodes can be encoded at the finest granularity $\textbf{Z} \in \mathbf{R}^{|V| \times |T| \times d}$ since all the nodes are known during learning, as shown in Fig. \textbf{\ref{fig:2_5_emb_stg}}. 
When the DTDG node set changes across snapshots in a transductive task (i.e. for \textit{case} (3) \cite{manessi2020dynamic,wang2020generic,behrouz2022anomaly}), the nodes can still be encoded in the shape of ${|V| \times |T| \times d}$, where $|V|$ denotes the cardinal of the universal node set \cite{wang2020generic}, by filling the missing values. An example of filling the missing values with \textbf{0} vectors \cite{manessi2020dynamic}  is shown in Fig. \ref{fig:2_5_emb} (A) for node C at $t_1$ and node A at $t_3$. 

For discrete time inductive settings (i.e. for \textit{cases} (4) $Ind_v$ \cite{pareja2020evolvegcn,hajiramezanali2019variational} and (5) $Ind_{DG}$ \cite{choi2021dynamic,sun2022ddgcn,wang2020fake}) the predictor cannot determine the existence of a node until it appears. This case is illustrated on Fig. \ref{fig:2_5_emb} (B) where the predictor cannot determine the existence of node C until it appears at $t_4$. Hence, $|V^t|$ can vary at each timestep $t$. Therefore, the embedding of nodes in the inference set cannot be represented in the shape of ${|V| \times |T| \times d}$. In this situation, one can use a list to store the representations of all accessible time steps for each node seen \cite{hajiramezanali2019variational}, e.g. $\mathbf{Z}=\left\{\mathbf{Z}^{1}, \mathbf{Z}^{2}, \ldots, \mathbf{Z}^{T}\right\}$ with $\mathbf{Z}^{t}\in
 \mathbb{R}^{|V|_t \times d}$ for $t \in \left\{1,\ldots,T\right\}$.

In continuous time, there is no longer a time grid, as shown in Fig. \ref{fig:2_5_emb} (C \& D). Therefore, there are no longer embedding updates for all nodes at each time step. Instead, when an event occurs on the CTDG, either the embeddings of the associated nodes are updated or the embedding(s) of the unseen node(s) are added \cite{rossi2020temporal,xu2020inductive}.

To informatively encode dynamic graphs into tensors or a list of vectors, a DGNN must capture both the structure information and their evolution over time. Therefore, to handle topology and time respectively, DGs are often decomposed or transformed into components like equivalent static (sub)graphs \cite{jain2016structural,yan2018spatial,ding2021simple}, random walks \cite{wang2021inductive,sato2019dyane,khoshraftar2019dynamic,nguyen2018continuous,lin2020t}, or sequences of matrices \cite{jia2020graphsleepnet,choi2021dynamic,manessi2020dynamic}. In the literature, numerous approaches have emerged by combining different encoders $f_G(\cdot)$ for static graphs with $f_T(\cdot)$ for temporal data. A plethora of graph and temporal data encoders have been at the root of the DG encoders reviewed in the section. These encoders are 
described in appendices B and C.

\begin{figure*}[!ht]
	\centering
		\includegraphics[width=\textwidth]{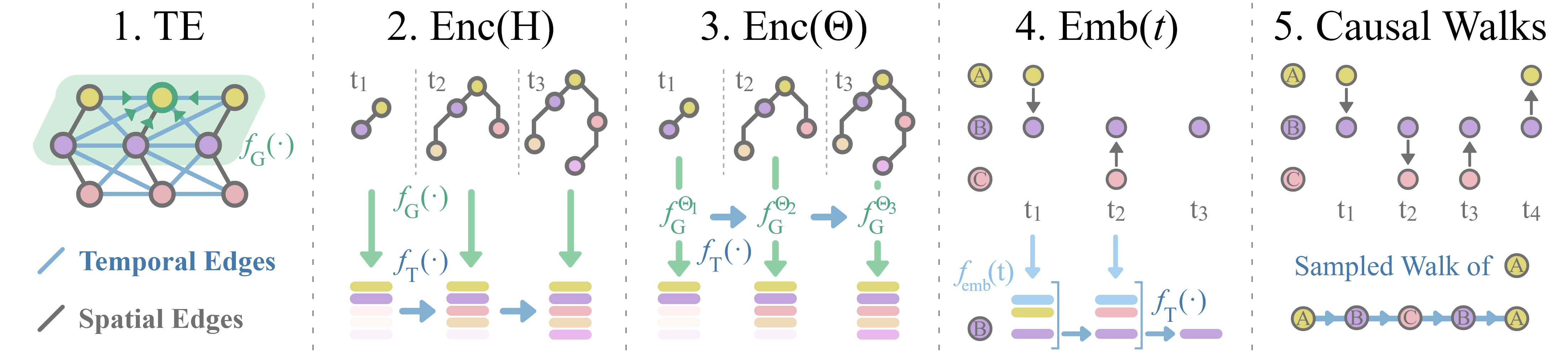}
	\caption{Dynamic graph neural network taxonomy on temporal information processing: 1. \textbf{Temporal Edge Modeling} models temporal edges to transform STGs into static graphs. 2. \textbf{Sequentially Encoding Hidden States H} encodes the hidden states of each snapshot across time with a temporal encoder $f_T(\cdot)$. 3. \textbf{Sequentially Encoding Parameters $\Theta$} encodes the parameters $\Theta$ of the graph encoder $f_G(\cdot)$ across time with a temporal encoder $f_T(\cdot)$. 4. \textbf{Embedding Time $t$} converts time values to vectors and concatenates or adds them to the attribute vectors when encoding node B. 5. \textbf{Causal Walks} restricts the random walks on dynamic graphs by causality.}
    \label{fig:DGNN_tax}
\end{figure*}

In the following subsections, we present a taxonomy of DGNN models which relies on five categories. Our categorization, illustrated in Figure \ref{fig:DGNN_tax}, is based on the strategy for handling both temporal and structural information.
\begin{enumerate}
    \item Modelling temporal edges and encoding \textit{ESG via topology}, denoted as $TE$ (Section 3.1).
    \item Sequentially encoding the hidden states, denoted as $enc(\textbf{H})$ (Section 3.2).
    \item Sequentially encoding the DGNN parameters, denoted as $enc(\Theta)$ (Section 3.3).
    \item Embedding occurrence time $t$ as edge feature of \textit{ESG via attributes}, denoted as $emb(t)$ (Section 3.4).
    \item Sampling causal walks, denoted as $Causal RW$ (Section 3.5).
\end{enumerate}

Note that these five approaches are not exclusive, i.e. they can be combined and used on the same DG. 

\begin{table*}[htb!]\centering
\caption{The available finest  embedding granularity under various dynamic graph learning cases.}\label{tab:2set}
\scriptsize
\begin{tabular}{|c|c|c|c|c|c|}\toprule 
DG Learning Case & (1) Trans-$fix_{V, E}$ & (2) Trans-$fix_{V}$ & (3) Trans-$vary$ & (4) Ind$_{V}$ & (5) Ind$_{DG}$ \\\cline{1-6}
Discrete Time  & \multicolumn{3}{c|}{(A)} & \multicolumn{2}{c|}{(B)}\\\cline{1-6}
Continuous Time  & \multicolumn{3}{c|}{(C)} & \multicolumn{2}{c|}{(D)}\\\cline{1-6}

\bottomrule
\end{tabular}
\end{table*}


\subsection{Temporal Edge Modeling}
Since applying convolution on a static graph is generally easier than encoding across multiple snapshots, the DG encoding problem is frequently transformed into encoding a static graph where each node is connected to itself in the adjacent snapshot \cite{jain2016structural,yan2018spatial}. This approach can also be interpreted as constructing a \textit{time-expanded graph} \cite{rehman2020exploring,azimifar2015vehicle,li2022enhanced} or \textit{node-oriented ESG} (see section 2.2.4) and is widely used to encode \textit{Trans-fix$_{V,E}$} cases, i.e. STGs. In more complex configurations, nodes are also connected with their k-hop neighbours in the adjacent snapshot(s) \cite{deng2020cola}. 
\begin{figure}[!ht]
	\centering
		\includegraphics[width=0.7\columnwidth]{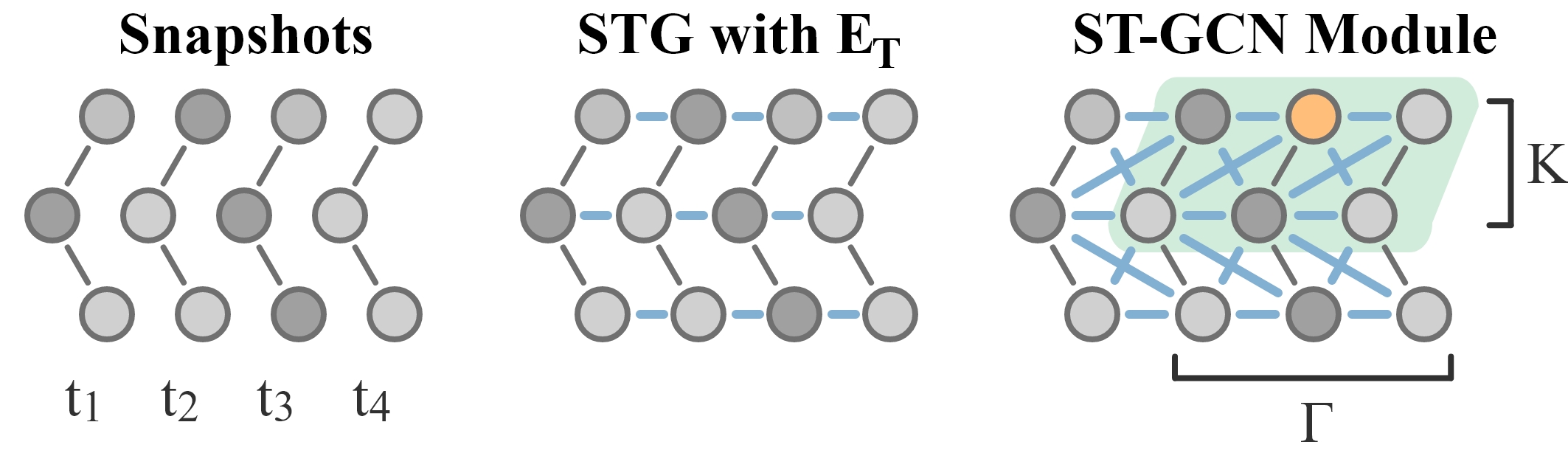}
	\caption{Comparing snapshot representation (left) and STG with temporal edges (middle) \cite{jain2016structural}. The edges in blue indicate temporal edges. In ST-GCN module (right) \cite{yan2018spatial}, temporal edges also connect nodes with their K-hop neighbours in the adjacent $\Gamma/2$ snapshot(s), thus the light green area shows the neighbourhood of the orange node while applying spatial-temporal graph convolution with K=1 and $\Gamma$=3.} 
	\label{FIG:STE}
\end{figure}

A simple example of such a strategy is shown in fig. \ref{FIG:STE}. An equivalent static graph $G’=\left\{V',E'_S,E'_T\right\}$ is obtained by modelling temporal edges \cite{jain2016structural}. $G’$ has $|V'|=|V|\times T$ nodes and $|E'_S|=|E|\times T$ spatial edges. Depending on the modelling approach, the number of temporal edges $|E'_T| = |V| \times (T-1)$ can be greater.

Once defined the connection rule for temporal edges, the traditional convolution for static graphs is applicable on ESGs. A state-of-the-art example is the \textbf{ST-GCN} module \cite{yan2018spatial}. To update the hidden states $h$ to $h'$ in an ST-GCN layer (see Eqn. \ref{equ:st-gcn_yan}), a typical spatial GNN structure aggregates the neighbour features with the msg($\cdot$) function, computes their weights by the w($\cdot$) function, and then sums them after normalisation with the norm($\cdot$) function. Note that, unlike the neighbourhood definition in static graphs, the neighbourhood set \textbf{N} of $v_i^t$ is defined by: (1) spatially, the shortest path distance with neighbour $v_j$ obeys $d(v_j,v_i) \leq K$ and (2) temporally, the time difference between timestamps $q$ and $t$, i.e. $|q-t|$ is not greater than $\lfloor\Gamma/2\rfloor$.

\begin{eqnarray}
    \label{equ:st-gcn_yan}
    h_i^{'t} &=& \sum_{v_j^t \in \textbf{N}_{v_i^t} } \textbf{norm}\left(\textbf{msg}(h_i^t, h_j^t) \cdot \textbf{w}(h_i^t,h_j^t)\right) \nonumber \\
    \textbf{N}_{v_{i}^t} &=& \left\{ {v_j^q | d(v_j^t,v_i^t) \leq K, |q-t| \leq \lfloor\Gamma/2\rfloor} \right\}
\end{eqnarray}


\subsection{Sequentially Encoding Hidden States H} \label{sec:3_ench}
In DTDGs, there are usually additions/deletions of edges or nodes \cite{singer2019node,li2019predicting,sato2019dyane,zhou2019dynamic}. To deal with these topological changes, this category denoted as $enc(\textbf{H})$ uses $f_G(\cdot)$ and $f_T(\cdot)$ to encode the graph and time domains alternatively. 
$Enc(\textbf{H})$ is widely applied to the Trans-$fix_{V,E}$ \textit{case},  i.e. STGs \cite{chaolong2018spatio,wu2019graph,yu2017spatio,li2017diffusion,ruiz2020gated,wu2020connecting,fan2022gnn} and the Trans-$fix_{V}$ \textit{cases} on DTDGs \cite{singer2019node,li2019predicting,sato2019dyane,zhou2019dynamic}. 
$f_T(\cdot)$ either encodes each snapshot across time after $f_G(\cdot)$ encodes each snapshot \cite{manessi2020dynamic,wang2020generic,seo2018structured}, i.e., in a stacked way \cite{skarding2021foundations}, or incorporates graph convolution when encoding each snapshot across time \cite{seo2018structured,chen2018gc,li2017diffusion}, i.e., in an integrated way \cite{skarding2021foundations}.

When the input is an STG, rather than processing them as static graphs, this approach factorises space and time and processes them differently \cite{yu2017spatio,wu2019graph,nicolicioiu2019recurrent,duta2020dynamic,fan2022gnn,kapoor2020examining,kim2021learning,jia2020graphsleepnet,li2017diffusion}. \textbf{RSTG} \cite{nicolicioiu2019recurrent} and \textbf{DyReG} \cite{duta2020dynamic} first encode each snapshot at node-level with weighted message passing as $f_G(\cdot)$, and then encode each node over time using LSTM or GRU as $f_T(\cdot)$. Both approaches are of the stacked fashion. 

\begin{table*}[!h]\centering
\caption{Main components of selected (DT)DGNNs which sequentially encode hidden states. Without specification, GNN refers to graph convolution or message passing, OP refers to Orthogonal Procrustes \cite{schonemann1966generalized}, TCN refers to temporal convolution and its variants, Attention refers to the attention mechanism \cite{bahdanau2014neural}, AE refers Autoencoder and its variants, and RW refers to Random Walk-based approaches \cite{tang2015line,perozzi2014deepwalk,grover2016node2vec}.}
\label{tab: ench}
\scriptsize
\begin{tabular}{lrrrrrr}\toprule
Model &Task case &Structure &Graph  &Temporal &Embedded &Ref. \\
 & & &Encoding &Encoding &Object &\\\midrule

DyReg &Trans-fix$_{V, E}$ &stacked &GNN &GRU &Node &\cite{duta2020dynamic}\\
RSTG  &Trans-fix$_{V, E}$ &stacked &GNN &LSTM &Node &\cite{nicolicioiu2019recurrent}\\
STAGIN &Trans-fix$_{V, E}$ &stacked &GIN &Transformer &Graph&\cite{kim2021learning} \\
GraphSleepNet &Trans-fix$_{V, E}$ &stacked &GNN &TCN &Graph&\cite{jia2020graphsleepnet}\\
GNN-RNN &Trans-fix$_{V, E}$ &stacked &GNN &LSTM &Node &\cite{fan2022gnn}\\
Graph WaveNet &Trans-fix$_{V, E}$ &stacked &GNN &TCN &Node &\cite{wu2019graph}\\
STGCN &Trans-fix$_{V, E}$ & stacked &GNN &TCN &Node &\cite{yu2017spatio}\\ 
DCRNN, GCRNN &Trans-fix$_{V, E}$ &integrated &GNN &GRU &Node &\cite{li2017diffusion}\\
USSTN &Trans-fix$_{V, E}$ &stacked &GNN &MLP &Node &\cite{kapoor2020examining}\\
GRNN &Trans-fix$_{V, E}$ &stacked & GNN & Gated RNN &Node &\cite{ruiz2020gated}\\
GCRN-M1 &Trans-fix$_{V, E}$ & stacked & GNN &LSTM &Node &\cite{seo2018structured} \\
GCRN-M2 &Trans-fix$_{V, E}$ & integrated & GNN &LSTM &Node &\cite{seo2018structured} \\

\hline

tNodeEmbed &Trans-fix$_{V}$ &stacked &RW &OP \& LSTM &Node &\cite{singer2019node}\\ 
DynSEM  &Trans-fix$_{V}$ &stacked &RW &OP &Node &\cite{zhou2019dynamic}\\ 
GC-LSTM &Trans-fix$_{V}$ &integrated & GNN &LSTM &Node &\cite{chen2018gc} \\
LRGCN &Trans-fix$_{V}$ &stacked & GNN &LSTM & Path &\cite{li2019predicting}\\ 
 & & & & & (set of edges) & \\ 
E-LSTM-D &Trans-fix$_{V}$ &stacked &AE &LSTM &Node &\cite{chen2019lstm}\\

\hline

WD-GCN, CD-GCN &Trans-vary &stacked & GNN &LSTM &Node &\cite{manessi2020dynamic}\\ 
TNDCN &Trans-vary &stacked & GNN & TCN &Node &\cite{wang2020generic}\\ 
ANOMULY &Trans-vary  & stacked & GNN & GRU & Node & \cite{behrouz2022anomaly}\\

\hline

Dyn-GCN &Ind$_{DG}$ &stacked & GNN &Attention &Graph&\cite{choi2021dynamic}\\
DDGCN &Ind$_{DG}$ & stacked & GNN &Temporal  &Graph&\cite{sun2022ddgcn}\\  
 & & & &Fusion & & \\  

\bottomrule
\end{tabular}
\end{table*}

If the node set $V_t$ of a DTDG is constant, then this DTDG is equivalent to an STG only with an additional change in the edge set $E_t$, i.e. Trans-${fix_{V}}$ \textit{case}. 
Therefore, the stacked architecture mentioned in the previous paragraph is equally practicable. 
A typical example is \textbf{CD-GCN} \cite{manessi2020dynamic} that concatenates the attributes and the adjacency matrix for each snapshot $t$ to form input $X^t || A^t$. Each snapshot is first encoded by GCN to obtain the node-level topological hidden states $\textbf{z}_{i,\textit{GCN}}^t$, and then encoded by LSTM in the time dimension to obtain $\textbf{z}_{i,\textit{LSTM}}^t$. Finally, an MLP maps the concatenation of the hidden states and raw features to the final node-level hidden states $\textbf{z}_{i}^t$ for each time step. Similar structures which stack temporal encoder $f_T(\cdot)$ after graph encoder $f_G(\cdot)$ are \textbf{STGCN} \cite{yu2017spatio}, \textbf{GraphSleepNet} \cite{jia2020graphsleepnet}, \textbf{E-LSTM-D} \cite{chen2019lstm}, \textbf{Graph WaveNet} \cite{wu2019graph}, \textbf{GRNN} \cite{ruiz2020gated}. A simple example is shown in Eqn. \ref{encH}, as a matter of fact, they can be stacked in more complex ways, see tab \ref{tab: ench} for more details.  

\begin{eqnarray} \label{encH}
    \textbf{Z}^t &=& f_T\left(f_G(\textbf{X}^t)\right) \mathrm{\ , or\ } f_T(\textbf{X}^t) \oplus f_G(\textbf{X}^t)
\end{eqnarray}

Another strategy for sequentially encoding the hidden states incorporates $f_G(\cdot)$ into $f_T(\cdot)$ rather than stacking them. Since there are usually projection or convolution modules in $f_T(\cdot)$ to handle the features of nodes, this "integrated mode" turns these modules into $f_G(\cdot)$ to aggregate neighbouring features. \textbf{GVRN-M2} \cite{seo2018structured} replaces the 2D convolution in convLSTM with graph convolution. Similar examples are \textbf{GC-LSTM} \cite{chen2018gc} and \textbf{DCRNN} \cite{li2017diffusion} which replace the linear layer in LSTM and GRU with respectively GCN\cite{kipf2016semi} and diffusion convolution \cite{teng2016scalable}.

To deal with the addition/deletion of nodes in a transductive nature, i.e. $Trans_{vary}$ \textit{case}, 
one needs to handle $|V_t|$ which may change across snapshots. \textbf{TNDCN} \cite{wang2020generic} proposes to set a universal node set $V = \cup V_t$ to ensure that $|V|$ is the same for each snapshot, which transforms the transductive \textit{case} into a node set-invariant \textit{case} on DTDG.

In the inductive case, one cannot presume $|V|$ to set the universal node set, which brings about an inconsistent number of nodes in each snapshot making $f_T(\cdot)$ impossible to encode at the node level. Therefore, this method in inductive tasks is only applicable for encoding graph-level representations across time, e.g. fake news detection based on its propagation tree. \textbf{Dyn-GCN} \cite{choi2021dynamic} applies Bi-GCN \cite{bian2020rumor} to encode the hidden states $\textbf{z}_{G_t}$ for each snapshot $t$ by aggregating the hidden states of its nodes and edges, and passes $(\textbf{z}_{G_1}, \textbf{z}_{G_2}, \ldots, \textbf{z}_{G_T})$ into an attention layer to compute the final hidden states of the entire DG.

\subsection{Sequentially Encoding Parameters $\Theta$}

Although $enc({\textbf{H}})$ is relatively intuitive and simple in terms of model structure, the problem is that they can neither handle the frequent changes of the node set, especially in the inductive tasks, nor pass learned parameters of $f_G(\cdot)$ across time steps \cite{pareja2020evolvegcn}. 
To encode DTDGs more flexibly, some other approaches constraint or encode the parameters of $f_G(\cdot)$ across time steps \cite{goyal2018dyngem,hajiramezanali2019variational,pareja2020evolvegcn}.

In order to encode the parameters $\Theta$ of GCN, \textbf{EvolveGCN} \cite{pareja2020evolvegcn} proposes to use LSTM or GRU to update the parameters of the GCN model at each time step as shown in eqn. \ref{eqn:EvolveGCN} and eqn. \ref{eqn:EvolveGCN_H}:
\begin{eqnarray} \label{eqn:EvolveGCN}
    \Theta_{f_G}^t &=& \mathrm{LSTM}(\Theta_{f_G}^{t-1})
\end{eqnarray}
\begin{eqnarray} \label{eqn:EvolveGCN_H}
    \Theta_{f_G}^t &=& \mathrm{GRU}(\textbf{H}^t, \Theta_{f_G}^{t-1})
\end{eqnarray}

Otherwise, by constraining the GNN parameters, \textbf{DynGEM} \cite{goyal2018dyngem} incorporates autoencoder (AE) to encode and reconstruct the adjacency matrix of each snapshot. Its parameters $\Theta_{t}$ are initialized with $\Theta_{t-1}$ to accelerate and stabilize the model training. Similarly, in \textbf{VGRNN} \cite{hajiramezanali2019variational} the authors combine GRNN and a variational graph AE in order to reuse learned hidden states of $t^-$ to compute the prior distribution parameters of the AE.

\subsection{Embedding Time $t$}

When the scale of the dynamic graph is large, as in social networks and recommendation systems, aggregation to snapshots is neither precise nor efficient \cite{skarding2021foundations,zaki2016comprehensive}. Its evolution is thus represented by a set of timestamped events. Therefore, when encoding a CTDG, one should consider not only how to asynchronously update the node representations over time, but also to define the neighbours of nodes. 

\begin{figure}[!h]
	\centering
    \includegraphics[width=0.6\columnwidth]{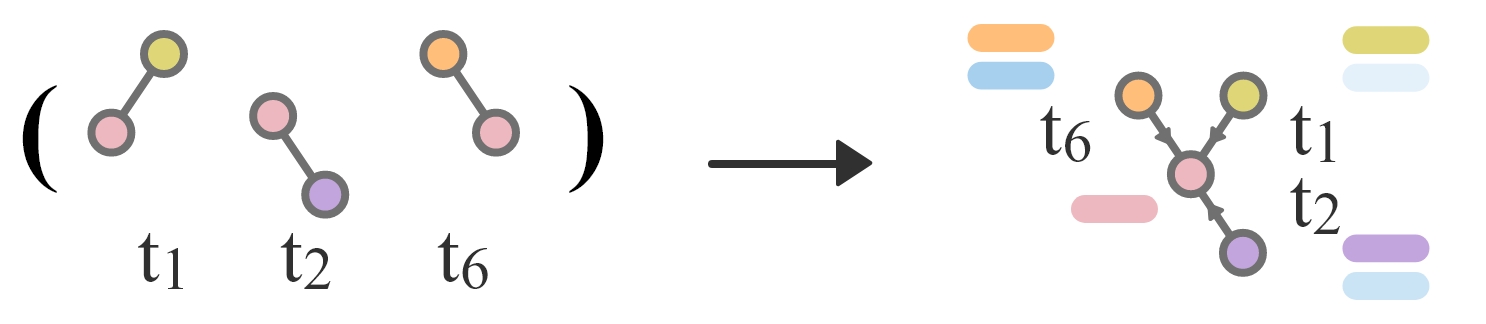}
	\caption{To aggregate information of neighbours on a dynamic graph, for example, to encode the pink node, the nodes with which it has an edge, i.e. its \textit{temporal neighbours \cite{ding2021simple}} (yellow, orange, purple nodes) and the temporal information (in blue) are embedded as vectors. This can also be interpreted as constructing a \textit{equivalent static graph} via attributes.}
	\label{fig:3_4_ctseg}
\end{figure}

The DGNNs presented in this subsection update the representation of a node $v$ when it changes, i.e. when it participates in a new edge or when its attributes change. Its neighbours at moment $t$, also called \textit{temporal  neighbourhood}, are usually defined as the nodes that have common edges with $v$ (before $t$) \cite{xu2020inductive}. Thus, such models consider $v$-centric equivalent static subgraph as shown in Fig. \ref{fig:3_4_ctseg}. In such cases, the occurrence time of each edge can be considered as part of the edge attributes or used as weight for graph convolution.

\textbf{TDGNN} \cite{qu2020continuous} proposes a weighted graph convolution by assuming that the earlier an edge is created, the more weight this edge will have in aggregation, as shown in Equ. \ref{TDGNN}. In more detail, the weight $\alpha_{u,v}^t$ of an edge $(u,v)$ at moment $t$ is calculated by the softmax of the existence time $t-t_{u,v}$ of the edge.

\begin{equation} \label{TDGNN}
    \alpha_{u,v}^t = 
    \frac{e^{t-t_{u,v}}}{\sum_{u \in N_t(v) \cup v} e^{t-t_{u,v}}}
\end{equation}

On the other hand, \textbf{TGAT} \cite{xu2020inductive} embeds the value of the existence time $t-t_{(u,v)}$ of edge $(u,v)$ as a vector of $d_T$ dimensions and then concatenates it to the node $v$'s hidden states $\textbf{z}_v^t$. When aggregating neighbour information for a node $v$, the weight of each neighbour is computed by multi-head attention (Eqn. \ref{equ:tgat}). Similar methods has also been applied for time embedding in Dynamic Graph Transformers \cite{wang2021tcl,chang2020continuous}.


\begin{eqnarray}
    \label{equ:tgat}
    \mathrm{q}(\mathrm{t})&=&[\mathrm{Z}(\mathrm{t})]_0 \mathrm{W}_{\mathrm{Q}} \nonumber \\
    \mathrm{K}(\mathrm{t})&=&[\mathrm{Z}(\mathrm{t})]_{1: \mathrm{N}} \mathrm{W}_{\mathrm{K}} \nonumber \\
    \alpha_{u,v}^t&=&\frac{\exp \left(q_v^T k_u\right)} {\sum_{u \in N_t(v)} \exp \left(q_v^T k_u\right)}
\end{eqnarray}

To better reuse the messages aggregated for each node by TGAT, \textbf{TGN} \cite{rossi2020temporal} adds a memory mechanism which memorizes the historical information for each node $v$ with a memory vector $\textbf{s}_v$. This memory is updated after each time $t$ a node $v$ aggregates its neighbours' information $\textbf{m}_v^t$.

\begin{equation}
    \textbf{s}_v^t = \mathrm{MLP}(\textbf{m}_v^t, \textbf{s}_v^{t-})
\end{equation}

Similar to TGN, \textbf{TGNF}\cite{song2021temporally} embeds nodes and time information via TGAT\cite{xu2020inductive}, and then updates the node's memory $\textbf{S}$ via the temporal memory module (TMM). In order to learn variational information better, it calculates the similarity $(\textbf{S}_t,\textbf{S}_{t^-})$ of the memory $\textbf{S}$ at $t$ and $t^-$ during training as part of the loss, called Time Difference Network (TDN).

Some other approaches consider the interaction predictions on the graph as a Temporal Point Problem (TPP \cite{lewis1972multivariate}) and simulate the conditional intensity function that describes the probability of the event. \textbf{DyRep} \cite{trivedi2019dyrep} encodes a strength matrix $\textbf{S} \in R^{|V| \times |V|}$ to simulate the intensity function of the interactions between each node pair and uses \textbf{S} as the weights for graph convolution. $\textbf{S}$ is initialized by the adjacency matrix $\textbf{A}$ and updated when an interaction $(u,v)$ occurs at time $t$. In this case, the embedding of each node involved is updated. For example, the update of $v$ is the sum of the three embedding components given by Eqn. \ref{equ:3_4_DyRep}: the latest embedding of $v$, the aggregation of embeddings of $u$'s neighbor nodes, and the time gap between $v$’s last update and this update.

\begin{eqnarray} \label{equ:3_4_DyRep}
    \textbf{z}_v^t = \sigma\left(\textbf{W}_1 \textbf{h}_{\mathrm{N(u)}}^{t^-}
     + \textbf{W}_2 \textbf{z}_{v}^{t^-} + \textbf{W}_3 (t-t^-)\right)
\end{eqnarray}

All of the models mentioned above embed timestamps or time-gaps as part of the features. However, models such as TGAT are unable to capture accurate changes in structure without node features \cite{wang2021inductive}. As discussed above, DyREP \cite{trivedi2019dyrep} solves this problem but is unable to perform the inductive task as it relies on the intensity matrix $\textbf{S}$. This raises another challenge: How to learn in an inductive task based only on the topology when the nodes have no features.

\subsection{Causal Random Walks}
Random walk-based approaches do not aggregate the neighbourhood information of nodes, but sample node sequences to capture the local structure. To incorporate time information to the random walks, different ways of defining "causal walks" on dynamic graphs are derived.

\begin{figure}[!h]
	\centering
		\includegraphics[width=.7\columnwidth]{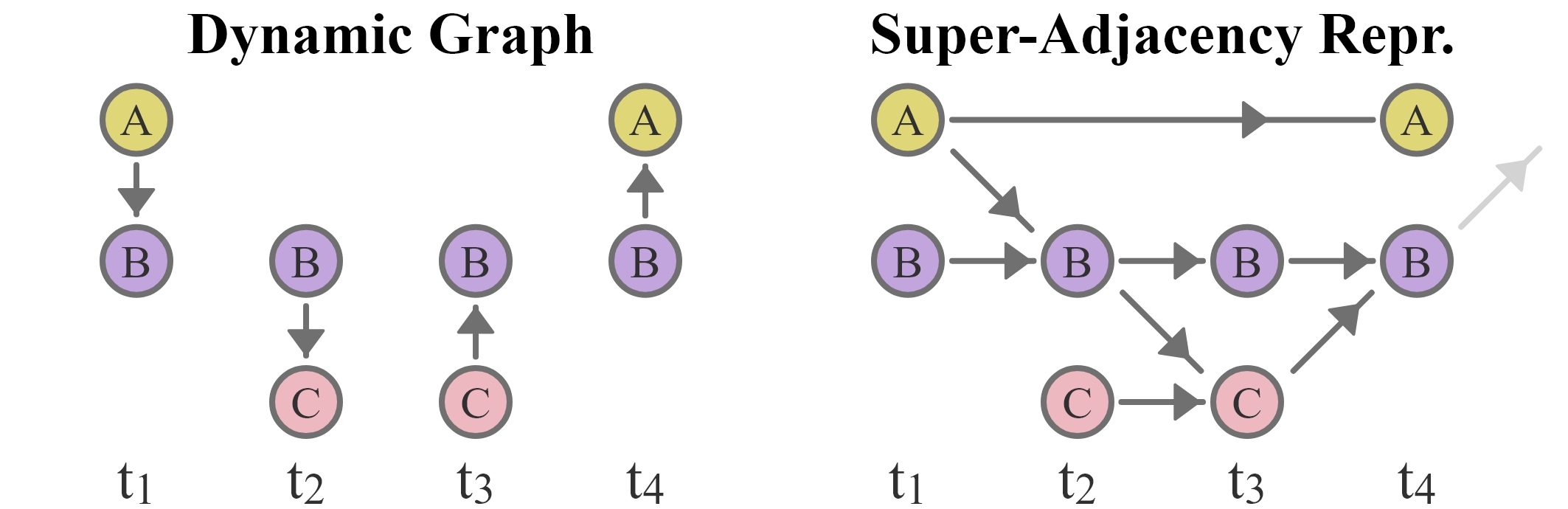}
	\caption{Supra-adjacency representation: treating discrete time dynamic graphs as directed static graphs for sampling random walks.} 
	\label{FIG:super-adja}
\end{figure}

To sample random walks with temporal information in discrete time, an intuitive way is to convert the DTDG into a directed static graph. Huang et al. \cite{huang2022learning} construct a \textit{ESG via topology}, and allow random walks to sample inside the current layer (/timestep) to get the structural information or walk into the previous layer to obtain historical evolving information. Following the supra-adjacency representation (see fig. \ref{FIG:super-adja}), \textbf{DyAne} \cite{sato2019dyane} applies the random walk approach of static graphs (DeepWalk \cite{perozzi2014deepwalk} and LINE \cite{tang2015line}) to DTDGs. Otherwise, without using these representations, \textbf{LSTM-node2vec} \cite{khoshraftar2019dynamic} directly samples one neighbouring node of a target node $v$ per snapshot as a walk.

\begin{figure}[!h]
	\centering
		\includegraphics[width=.8\columnwidth]{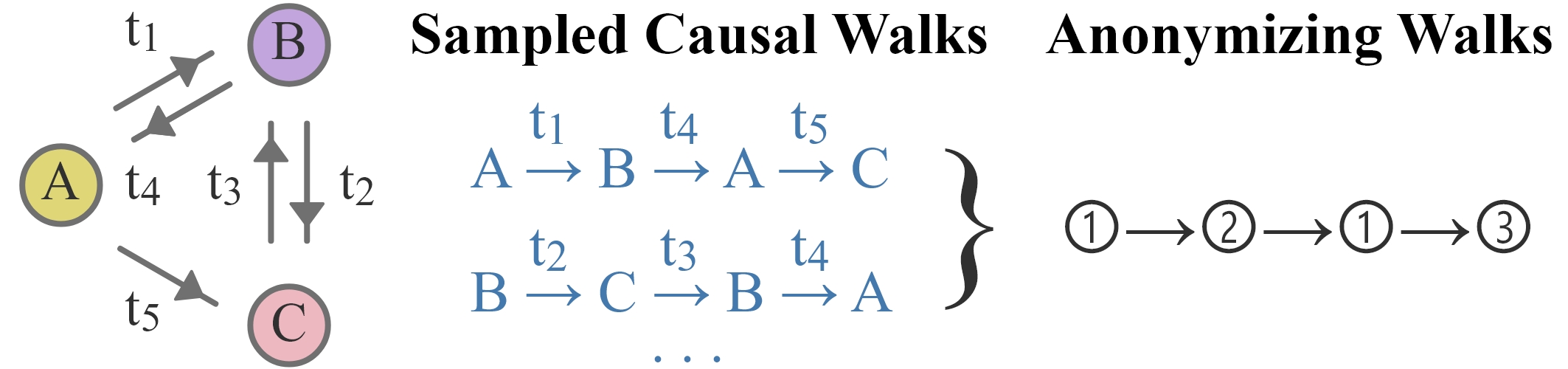}
	\caption{Schematic diagram on sampling casual walks on a CTDG, the right part shows graph pattern extracted by anonymizing nodes.} 
	\label{FIG:ct-walks}
\end{figure}

Similar to the method based on supra-adjacency, the walks obey causality under continuous-time, i.e. after walking from node B to node C via an edge occurring at $t2$, node C can only walk to the next node via an edge having occurring time $t > t2$ (Fig. \ref{FIG:ct-walks}). 

Some early methods are \textbf{CTDNE} \cite{nguyen2018continuous} and \textbf{T-Edge} \cite{lin2020t}. To encode a node $v$, they both sample the causal walks starting from $v$, each walk $W$ being a sequence of (node, time) pairs. Then they embed each (node, time) pair and encode the sequence $W$ by $f_T(\cdot)$ like RNNs. Combined with anonymous walk embedding, Causal Anonymous Walks (\textbf{CAW} \cite{wang2021inductive}) anonymise nodes and their attributes in order to focus more on graph motifs, which solve the problem at the end of Sec. 3.5.


\subsection{Encoding Heterogeneous Graphs}
\begin{figure}[!h]
	\centering
		\includegraphics[width=.8\columnwidth]{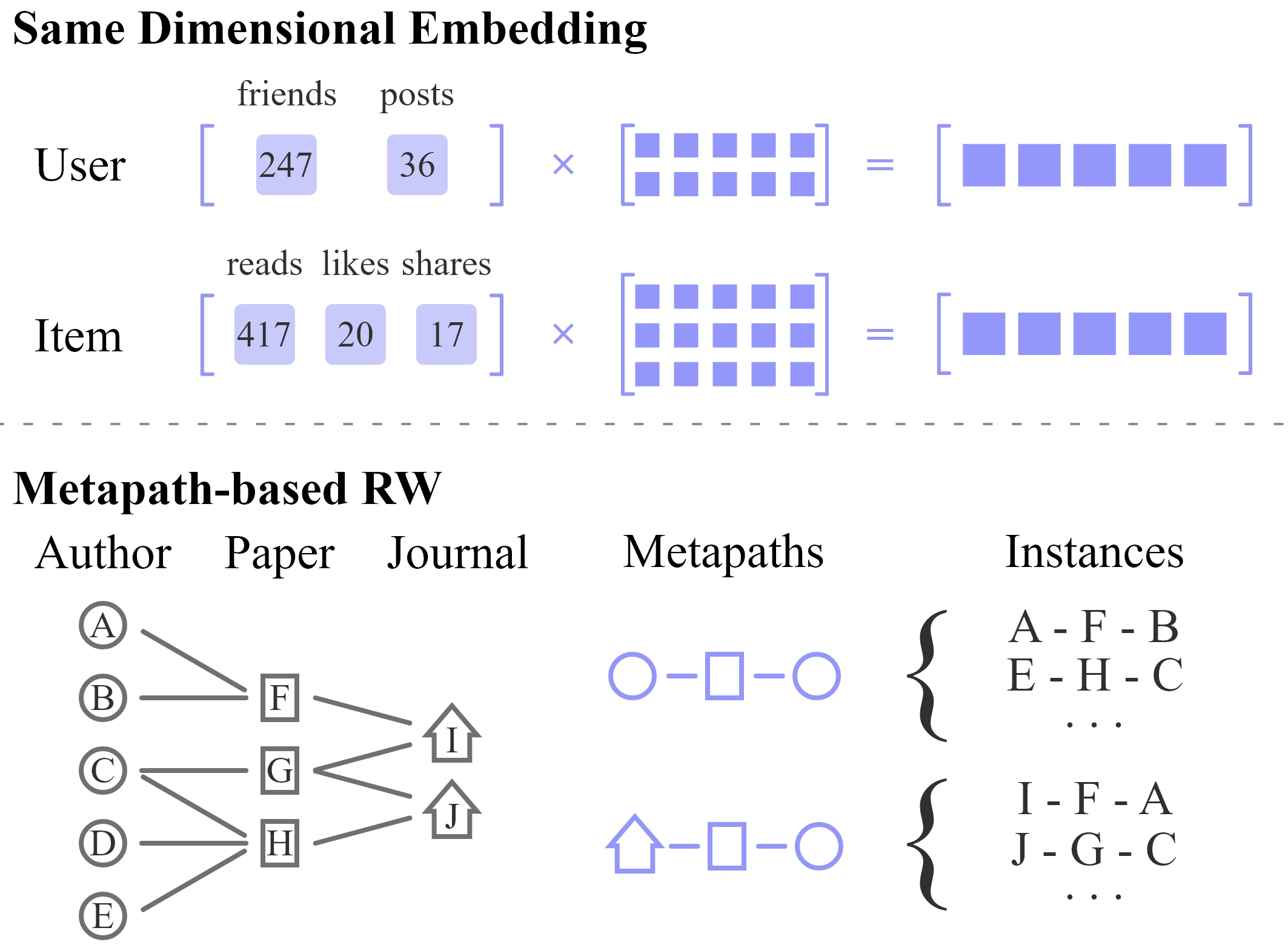}
	\caption{Handling heterogeneous nodes. Above: embedding them into the same vector space (e.g. with $d=5$). Below: sampling random walks with defined metapaths.} 
	\label{FIG:het}
\end{figure}

Since heterogeneous graphs can maintain more informative representations in tasks like link prediction in recommendation systems, a key challenge is to have a dedicated module for handling different types of nodes and edges on dynamic graphs. We introduce in the following paragraphs two main approaches as illustrated in Fig. \ref{FIG:het}.

GNN-based models keep the same dimension $d$ of embedding for different node types (e.g. for user nodes and item nodes) to facilitate computations. 
For example, to update the node embeddings when user $u$ and item $i$ have a connection $e_{u,i}^t$ at time $t$, \textbf{JODIE} \cite{kumar2019predicting} embeds the user attributes $\textbf{x}_u$, the item attributes $\textbf{x}_i$, the edge attributes $\textbf{x}_e$ and the time difference since the last update $\Delta$ to the same dimension $d$ and then adds them together : 

\begin{eqnarray}\label{equ_jodie}
    \textbf{h}_u^t = \sigma\left(\textbf{W}_1^u \textbf{h}_u^{t^-} + \textbf{W}_2^u \textbf{h}_i^{t^-} + \textbf{W}_3^u \textbf{h}_e^{t} + \textbf{W}_4^u \Delta_u\right) \nonumber\\ 
    \textbf{h}_i^t = \sigma\left(\textbf{W}_1^i \textbf{h}_i^{t^-} + \textbf{W}_2^i \textbf{h}_u^{t^-} + \textbf{W}_3^i \textbf{h}_e^{t} + \textbf{W}_4^i \Delta_i\right)
\end{eqnarray}
where $t^-$ refers to the timestamp of the last update thus $\Delta=t-t^-$, and $h_i^t=x_i^t$ for the first layer of model. \textbf{DMGCF}\cite{tang2021dynamic} and \textbf{DGCF}\cite{li2020dynamic} also use similar approaches for same dimensional embedding.

RW-based methods deal with heterogeneous graphs by defining metapaths \cite{dong2017metapath2vec} which specifies the type of each node in the walk, such as (user, item, user), so that each random walk sampled (also called "instance") with the same metapath can be projected to the same vector space. Examples on the dynamic graph are \textbf{THINE} \cite{huang2021temporal} and \textbf{HDGNN} \cite{zhou2020heterogeneous} which sample instances and encode them through the attention layer and bidirectional RNN layers, respectively.

\vspace{1em}

So far we have presented a wide variety of approaches to capture time and graph dependencies within dynamic graphs. In the next section, we discuss their global success and limitations, as well as present guidelines for the design of DGNN architectures.


\section{Guidelines for Designing DGNNs}
In the previous sections, we have highlighted the diversity of contexts which can be encountered when considering machine learning on dynamic graphs, as well as the diversity of existing models to tackle these problems. In this section, we present some guidelines for designing DGNNs on the basis of the taxonomies presented in section \ref{sec:2} and \ref{sec:DynamicGraphNeuralNetworks}. We also discuss the latest trends for optimizing DGNNs. To the best of our knowledge, while such guidelines have already been proposed for static graphs \cite{zhou2020graph},  they do not exist for DGNNs.  


\subsection{General Design Workflow of DGNNs}

For static graphs, Zhou et \emph{al.} \cite{zhou2020graph} described the GNN design pipeline as: \emph{i)} determine the input graph structure and scale, \emph{ii)} determine the output representation according to the downstream task, and \emph{iii)} add computational modules.

For dynamic graphs, the design of DGNN has to consider more factors. We therefore generalize the workflow of designing DGNN as follows :
\begin{enumerate}
    \item Clearly define the input, output and nature of the task, according to the taxonomies of section \ref{sec:2} ;
    \item Choose the compatible time encoding approach according to the learning setting, using the categorizations of section \ref{sec:DynamicGraphNeuralNetworks} and more precisely the Table \ref{tab: adapt} ;
    \item Design NN structure ;

    \item Optimise the DGNN model.
\end{enumerate}

The key points in the DGNN compatibility are the input time granularity, the nature, and the object to be encoded. We concluded their known adapted DG types and listed them in table \ref{tab: adapt} and describe the various cases in this subsection. 

\begin{table}[h!]\centering
\caption{Adapted dynamic graph types of each method: \\\textbf{TE} stands for "Temporal Edge Modeling", \textbf{Enc(H)} for "Sequentially Encoding Hidden States", \textbf{Enc($\Theta$)} for "Sequentially Encoding Model Parameters", \textbf{Emb(t)} for "Embedding time", \textbf{Causal RW} for "Causal Random Walks". $\checkmark$ means "applicable", $\ast$ means "applicable with output restrictions", and no symbol indicates that it is not yet used in the literature.}\label{tab: adapt}

\begin{tabular}{lccc}\toprule
Approach &DT Trans. &DT Ind. &CT Trans.\&Ind. \\\midrule
\ding{172} TE & \checkmark & & \\
\ding{173} Enc(H) & \checkmark & $\ast$ & \\
\ding{174} Enc($\Theta$) & \checkmark & \checkmark & \\
\ding{175} Emb(t) & \checkmark & \checkmark & \checkmark \\
\ding{176} Causal RW & \checkmark & \checkmark & \checkmark \\
\bottomrule
\end{tabular}
\end{table}



Transductive tasks under discrete time, as a relatively simple setting, can be encoded with any approach to incorporate time information. In the setting of inductive DT, no approach using \textit{TE} is found in the literature, and the method of $enc(\textbf{H})$ must also have its output aggregated because of the gap mentioned in sections \ref{sec:3_ench}, e.g. node-level time-aggregated or graph-level time-step. To handle continuous time, only \textit{emb(t)} and \textit{Causal RW} are widely used in the literature.

Once the approach has been selected, the next step is to add the computational components, which are very different for each approach.

For the first three methods mainly used in discrete time, i.e., \ding{172} $TE$, \ding{173} $Enc(\textbf{H})$, and \ding{174} $Enc(\Theta)$, they can be abstracted to encode graph and time information via $f_G(\cdot)$ and $f_T(\cdot)$, respectively. These neural network components are usually some of the modules introduced in the appendices B, C and Tab. \ref{tab: ench}. In particular, the encoding of temporal edges in the $TE$ approach can be performed by $f_G(\cdot)$ without a specific $f_T(\cdot)$. In $enc(\textbf{H})$ approach, there are more possible ways to combine $f_G(\cdot)$ and $f_T(\cdot)$ (stacked, integrated, etc.). In $enc(\Theta)$ approach, the key is how $f_T(\cdot)$ acts on the parameters of $f_G(\cdot)$. 

For the time embedding approach \ding{175} \textit{emb(t)}, in addition to the method of aggregating information of neighbours i.e. $f_G(\cdot)$, one also needs to consider the function for time embedding (e.g. by a set of sine or cosine functions \cite{xu2020inductive} or a learnable linear layer \cite{trivedi2019dyrep}) and how to combine them to the hidden states of edges or nodes (e.g. by addition or concatenation). In the situation where the past representation of a node is stored \cite{song2021temporally,rossi2020temporal,trivedi2019dyrep}, one also needs to determine the module used to update the node representation, i.e. $f_T(\cdot)$.

For random walk-based method \ding{176} \textit{Causal RW}, one needs to determine the random walk  sampling strategy as well as the method for encoding the sampled walks, typical walk-encoding strategies are \cite{grover2016node2vec,perozzi2014deepwalk,tang2015line,wang2016structural}.

Last but not least, in the case of heterogeneous graphs, an additional vector projection or setting of metapaths has to be considered.

\subsection{Trends in Optimising DGNNs}
With the development of artificial neural networks and the continuous emergence of new structures over the last five years, questions have been raised about the optimisation of DGNNs. Apart from general neural network training issues (overfitting, lack of data, vanishing gradient, etc.), a main issue for GNN is over-smoothing \cite{zhao2019pairnorm,chen2020measuring,cai2020note}: if the number of layers and iterations of a GNN is too large, the hidden states of each node will converge to the same value. The second main challenge is over-squashing \cite{alon2020bottleneck,topping2021understanding,di2023over}: if a node has a very large number of K-hop neighbours, then the information passed from a distant node will be compressed and distorted. 

To overcome the above problems, numerous methods have been proposed. According to us, general trends to improve DGNN can be categorised as (1) Input oriented, (2) DGNN component oriented, and (3) DGNN structure oriented.

\subsubsection{Input Oriented Optimisation}
To avoid overfitting when learning graph representations, two main issues related to the DG need to be considered: the noise in topology (e.g. missing or incorrect links) and the noise in the attributes (e.g. incorrect input attributes or output labels) \cite{kong2020flag,zhao2021data}.

Since there may be pairs of similar nodes in the graph that should be connected but are not (due to geographical constraints, missing data, etc.), some approaches aim to exploit more informative graph structure or to augment attribute data.

For discrete time, an example is \textbf{ST-SHN} \cite{xia2022spatial} for crime prediction. Considering each region as a node and its geographical connectivity as an edge, ST-SHN infers the hyper-edges connecting multiple regions by learning the similarity of hidden states between node pairs. These hyper-edges help to learn cross-region relations to handle the global context.

For continuous time, such as heterogeneous graph-based recommendation systems, \textbf{DMGCF} \cite{tang2021dynamic} constructs two additional homogeneous graphs $G_u$ for users and $G_{i}$ for items based on the known user-item graph $G_{ui}$. Then two GCNs are used to aggregate information on $G_{u}\cup G_{i}$ and $G_{ui}$ respectively to learn more informative node embeddings.

Noise in attributes can also lead to overfitting DGNNs, therefore adaptive data augmentation is another direction for input-oriented improvement. Wang \emph{et al.} proposed Memory Tower Augmentation (\textbf{MeTA} \cite{wang2021adaptive}) for continuous time data augmentation by perturbing time, removing edges, and adding edges with perturbed time. Each augmentation has learnable parameters to better adapt to different input data.

\subsubsection{DGNN Component Oriented Optimisation}
To solve the over-smoothing and over-squashing problems on graphs, the improvement of the DGNN modules focuses on a more versatile message propagation and a more efficient aggregation.

To avoid stacking multiple layers of GCNs, \textbf{TNDCN} \cite{wang2020generic} uses different-step network diffusion which provides a larger receptive field for each layer. Each step $k$ propagates attributes with a $k$-hop neighbourhood and has its independent learnable parameters $\mathrm{\Gamma}_k$, as shown in equ. \ref{equ_TNDCN}, where $\mathrm{\widetilde{A}^k}$ refers to the parameterised $k$-hop connectivity matrix.

\begin{eqnarray} \label{equ_TNDCN}
    \mathrm{\textbf{H}} = \sum_{k\geq0} \mathrm{\widetilde{\textbf{A}}^k} \mathrm{\textbf{H}}^k \mathrm{\Gamma}_k
\end{eqnarray}

 Inspired from the Bidirectionnal LSTM \cite{blstm2013}, an easy-to-implement enhancement for propagation is bi-directional message passing \textbf{Bi-GCN} \cite{bian2020rumor,choi2021dynamic}. It processes an undirected tree graph as two directed tree graphs: the first one from the root to the leaves, and the second one from leaves to the root. Then two GCNs with different parameters encode each case independently to obtain more informative hidden states. 

Besides improving propagation methods, a widely used technique for aggregating information is the (multi-head) attention mechanism, which enables the model to use adaptive weights $\left(f_{attn}: R^{|N_{v_i}| \times d} \to R^{|N_{v_i}| \times 1}\right)$ for enhancing vanilla GCNs \cite{tiukhova2022influencer}, as well as its variant self-attention mechanism $\left(f_{self-attn}: R^{|N_{v_i}| \times d} \to R^{|N_{v_i}| \times d'}\right)$ which is good at capturing the internal correlation among elements in the sequence.

The three component-oriented improvements mentioned above are applicable not only to $f_G(\cdot)$, but also to $f_T(\cdot)$ when encoding along time, such as dilated convolution for expanding the temporal perceptual field \cite{dilated}, Bi-LSTM for bidirectional propagation along the sequence \cite{blstm2013}, and (self-)attention mechanisms for encoding sequences \cite{vaswani2017attention}.

\subsubsection{DGNN Structure Oriented Optimisation}
Another research direction is to optimise the overall structure. For example, 
residual connection \cite{he2015deep} for dealing with the vanishing gradient is widely used in DGNNs, especially for the DGNNs which sequentially encode hidden states \cite{kapoor2020examining,manessi2020dynamic,li2017diffusion}.

\begin{figure*}[!ht]
    \centering		\includegraphics[width=.85\textwidth]{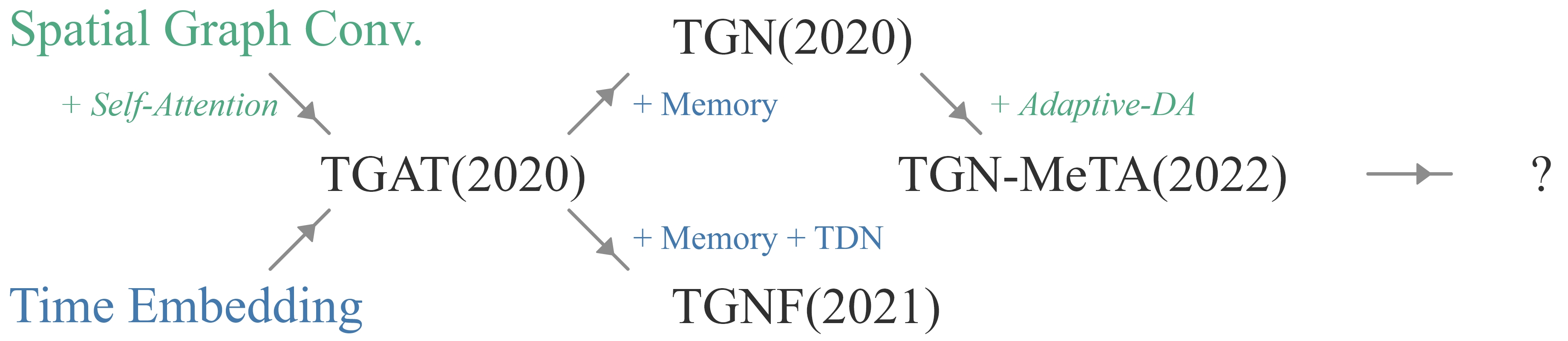}
	\caption{The main differences between the time-embedding methods, with the green color indicating the graph encoding-related methods and the blue color indicating the time encoding-related methods.} 
	\label{FIG:CTDGNN}
\end{figure*}

In particular for the time embedding approach under continuous time, how to better store historical information and how to learn variations of information are two major challenges due to the difficulty of encoding over time. Models such as \textbf{TGAT} \cite{xu2020inductive} can encode the timestamps of the event but cannot reuse the hidden states already encoded in the past timesteps. \textbf{TGN} \cite{rossi2020temporal} addresses this problem by incorporating memory modules, \textbf{TGNF} \cite{song2021temporally} adds the similarity between current memory and previous memory in the loss to encourage the model to learn variational information. All these improvements enhance the performance and efficiency of the model. The evolution of the relevant models is shown in Fig. \ref{FIG:CTDGNN}.

\section{Conclusion}

Thanks to their ability to integrate both structural and temporal aspects in a compact formalism, dynamic graphs has emerged in the last few years as a state of the art model for describing dynamic systems. 

Many scientific communities have investigated this area of research using their own definition, their own vocabulary, their own constraints, and have proposed prediction models dedicated to their downstream tasks. Among these models, Dynamic Graph Neural Networks occupy an important place, taking benefit of the representation learning paradigm.   

The first part of this survey provides a clarification and a categorization of the different dynamic graph learning contexts that are encountered in the literature, from the input point of view, the output point of view and the learning setting (inductive/transductive nature of the task). This categorization leads to five different learning \textit{cases} covering the different contexts encountered in the literature. 

Using this categorization, our second contribution was to propose a taxonomy of existing DGNN models. We distinguish five families of DGNN, according to the strategy used to incorporate time information in the model

Finally, we also provide to practitioners some guidelines for designing and improving DGNNs.

Although dynamic graph learning is a recent discipline, it will undoubtedly be a major trend for machine learning researchers for years to come. We hope that this review is a modest contribution in this direction.
\section{Acknowledgement}

This work was financially supported by the ANR Labcom Lisa ANR-20-LCV1-0009

\newpage
\appendix



\begin{thebibliography}{00}

\bibitem{wang2019heterogeneous}Wang, X., Ji, H., Shi, C., Wang, B., Ye, Y., Cui, P. \& Yu, P. Heterogeneous graph attention network. {\em The World Wide Web Conference}. pp. 2022-2032 (2019)
\bibitem{luo2020dynamic}Luo, W., Zhang, H., Yang, X., Bo, L., Yang, X., Li, Z., Qie, X. \& Ye, J. Dynamic heterogeneous graph neural network for real-time event prediction. {\em Proceedings Of The 26th ACM SIGKDD International Conference On Knowledge Discovery \& Data Mining}. pp. 3213-3223 (2020)
\bibitem{shi2022heterogeneous}Shi, C. Heterogeneous Graph Neural Networks. {\em Graph Neural Networks: Foundations, Frontiers, And Applications}. pp. 351-369 (2022)
\bibitem{hamilton2017representation}Hamilton, W., Ying, R. \& Leskovec, J. Representation learning on graphs: Methods and applications. {\em ArXiv Preprint ArXiv:1709.05584}. (2017)
\bibitem{cai2018comprehensive}Cai, H., Zheng, V. \& Chang, K. A comprehensive survey of graph embedding: Problems, techniques, and applications. {\em IEEE Transactions On Knowledge And Data Engineering}. \textbf{30}, 1616-1637 (2018)
\bibitem{skarding2021foundations}Skarding, J., Gabrys, B. \& Musial, K. Foundations and Modeling of Dynamic Networks Using Dynamic Graph Neural Networks: A Survey. {\em IEEE Access}. \textbf{9} pp. 79143-79168 (2021)
\bibitem{xue2022dynamic}Xue, G., Zhong, M., Li, J., Chen, J., Zhai, C. \& Kong, R. Dynamic network embedding survey. {\em Neurocomputing}. \textbf{472} pp. 212-223 (2022)
\bibitem{holme2012temporal}Holme, P. \& Saramäki, J. Temporal networks. {\em Physics Reports}. \textbf{519}, 97-125 (2012)
\bibitem{zaki2016comprehensive}Zaki, A., Attia, M., Hegazy, D. \& Amin, S. Comprehensive survey on dynamic graph models. {\em International Journal Of Advanced Computer Science And Applications}. \textbf{7} (2016)
\bibitem{zhang2010survey}Zhang, J. A survey on streaming algorithms for massive graphs. {\em Managing And Mining Graph Data}. pp. 393-420 (2010)
\bibitem{mcgregor2014graph}McGregor, A. Graph stream algorithms: a survey. {\em ACM SIGMOD Record}. \textbf{43}, 9-20 (2014)
\bibitem{kazemi2020representation}Kazemi, S., Goel, R., Jain, K., Kobyzev, I., Sethi, A., Forsyth, P. \& Poupart, P. Representation learning for dynamic graphs: A survey.. {\em J. Mach. Learn. Res.}. \textbf{21}, 1-73 (2020)
\bibitem{barros2021survey}Barros, C., Mendonça, M., Vieira, A. \& Ziviani, A. A survey on embedding dynamic graphs. {\em ACM Computing Surveys (CSUR)}. \textbf{55}, 1-37 (2021)
\bibitem{bronstein2017geometric}Bronstein, M., Bruna, J., LeCun, Y., Szlam, A. \& Vandergheynst, P. Geometric deep learning: going beyond euclidean data. {\em IEEE Signal Processing Magazine}. \textbf{34}, 18-42 (2017)
\bibitem{holme2015modern}Holme, P. Modern temporal network theory: a colloquium. {\em The European Physical Journal B}. \textbf{88} pp. 1-30 (2015)
\bibitem{guo2022continuous}Guo, J., Han, Z., Su, Z., Li, J., Tresp, V. \& Wang, Y. Continuous Temporal Graph Networks for Event-Based Graph Data. {\em ArXiv Preprint ArXiv:2205.15924}. (2022)
\bibitem{yang2008non}Yang, J., Yang, S., Fu, Y., Li, X. \& Huang, T. Non-negative graph embedding. {\em 2008 IEEE Conference On Computer Vision And Pattern Recognition}. pp. 1-8 (2008)
\bibitem{sarkar2011community}Sarkar, S. \& Dong, A. Community detection in graphs using singular value decomposition. {\em Physical Review E}. \textbf{83}, 046114 (2011)
\bibitem{jeh2002simrank}Jeh, G. \& Widom, J. Simrank: a measure of structural-context similarity. {\em Proceedings Of The Eighth ACM SIGKDD International Conference On Knowledge Discovery And Data Mining}. pp. 538-543 (2002)
\bibitem{page1999pagerank}Page, L., Brin, S., Motwani, R. \& Winograd, T. The PageRank citation ranking: Bringing order to the web.. (Stanford InfoLab,1999)
\bibitem{tong2006fast}Tong, H., Faloutsos, C. \& Pan, J. Fast random walk with restart and its applications. {\em Sixth International Conference On Data Mining (ICDM'06)}. pp. 613-622 (2006)
\bibitem{pearson1905problem}Pearson, K. The problem of the random walk. {\em Nature}. \textbf{72}, 294-294 (1905)
\bibitem{DBLP:journals/corr/Michail15}Michail, O. An Introduction to Temporal Graphs: An Algorithmic Perspective. {\em CoRR}. \textbf{abs/1503.00278} (2015), http://arxiv.org/abs/1503.00278
\bibitem{lewis1972multivariate}Lewis, P. Multivariate point processes. {\em Proceedings Of The Berkeley Symposium On Mathematical Statistics And Probability}. \textbf{1} pp. 401 (1972)
\bibitem{teng2016scalable}Teng, S. \& Others Scalable algorithms for data and network analysis. {\em Foundations And Trends® In Theoretical Computer Science}. \textbf{12}, 1-274 (2016)
\bibitem{schonemann1966generalized}Schönemann, P. A generalized solution of the orthogonal procrustes problem. {\em Psychometrika}. \textbf{31}, 1-10 (1966)
\bibitem{lecun1995convolutional}LeCun, Y., Bengio, Y. \& Others Convolutional networks for images, speech, and time series. {\em The Handbook Of Brain Theory And Neural Networks}. \textbf{3361}, 1995 (1995)
\bibitem{hopfield1982neural}Hopfield, J. Neural networks and physical systems with emergent collective computational abilities.. {\em Proceedings Of The National Academy Of Sciences}. \textbf{79}, 2554-2558 (1982)
\bibitem{hochreiter1997long}Hochreiter, S. \& Schmidhuber, J. Long short-term memory. {\em Neural Computation}. \textbf{9}, 1735-1780 (1997)
\bibitem{gers2001long}Gers, F. Long short-term memory in recurrent neural networks. (Verlag nicht ermittelbar,2001)
\bibitem{chung2014empirical}Chung, J., Gulcehre, C., Cho, K. \& Bengio, Y. Empirical evaluation of gated recurrent neural networks on sequence modeling. {\em ArXiv Preprint ArXiv:1412.3555}. (2014)
\bibitem{oord2016wavenet}Oord, A., Dieleman, S., Zen, H., Simonyan, K., Vinyals, O., Graves, A., Kalchbrenner, N., Senior, A. \& Kavukcuoglu, K. Wavenet: A generative model for raw audio. {\em ArXiv Preprint ArXiv:1609.03499}. (2016)
\bibitem{yu2015multi}Yu, F. \& Koltun, V. Multi-scale context aggregation by dilated convolutions. {\em ArXiv Preprint ArXiv:1511.07122}. (2015)
\bibitem{zhang2019multi}Zhang, Y., Thorburn, P. \& Fitch, P. Multi-task temporal convolutional network for predicting water quality sensor data. {\em International Conference On Neural Information Processing}. pp. 122-130 (2019)
\bibitem{bahdanau2014neural}Bahdanau, D., Cho, K. \& Bengio, Y. Neural machine translation by jointly learning to align and translate. {\em ArXiv Preprint ArXiv:1409.0473}. (2014)
\bibitem{vaswani2017attention}Vaswani, A., Shazeer, N., Parmar, N., Uszkoreit, J., Jones, L., Gomez, A., Kaiser, Ł. \& Polosukhin, I. Attention is all you need. {\em Advances In Neural Information Processing Systems}. \textbf{30} (2017)
\bibitem{loomis2013introduction}Loomis, L. Introduction to abstract harmonic analysis. (Courier Corporation,2013)
\bibitem{kazemi2019time2vec}Kazemi, S., Goel, R., Eghbali, S., Ramanan, J., Sahota, J., Thakur, S., Wu, S., Smyth, C., Poupart, P. \& Brubaker, M. Time2vec: Learning a vector representation of time. {\em ArXiv Preprint ArXiv:1907.05321}. (2019)
\bibitem{bruna2013spectral}Bruna, J., Zaremba, W., Szlam, A. \& LeCun, Y. Spectral networks and locally connected networks on graphs. {\em ArXiv Preprint ArXiv:1312.6203}. (2013)
\bibitem{defferrard2016convolutional}Defferrard, M., Bresson, X. \& Vandergheynst, P. Convolutional neural networks on graphs with fast localized spectral filtering. {\em Advances In Neural Information Processing Systems}. \textbf{29} (2016)
\bibitem{kipf2016semi}Kipf, T. \& Welling, M. Semi-supervised classification with graph convolutional networks. {\em ArXiv Preprint ArXiv:1609.02907}. (2016)
\bibitem{micheli2009neural}Micheli, A. Neural network for graphs: A contextual constructive approach. {\em IEEE Transactions On Neural Networks}. \textbf{20}, 498-511 (2009)
\bibitem{gilmer2017neural}Gilmer, J., Schoenholz, S., Riley, P., Vinyals, O. \& Dahl, G. Neural message passing for quantum chemistry. {\em International Conference On Machine Learning}. pp. 1263-1272 (2017)
\bibitem{li2015gated}Li, Y., Tarlow, D., Brockschmidt, M. \& Zemel, R. Gated graph sequence neural networks. {\em ArXiv Preprint ArXiv:1511.05493}. (2015)
\bibitem{dauphin2017language}Dauphin, Y., Fan, A., Auli, M. \& Grangier, D. Language modeling with gated convolutional networks. {\em International Conference On Machine Learning}. pp. 933-941 (2017)
\bibitem{hamilton2017inductive}Hamilton, W., Ying, Z. \& Leskovec, J. Inductive representation learning on large graphs. {\em Advances In Neural Information Processing Systems}. \textbf{30} (2017)
\bibitem{kipf2016variational}Kipf, T. \& Welling, M. Variational graph auto-encoders. {\em ArXiv Preprint ArXiv:1611.07308}. (2016)
\bibitem{lin2020graph}Lin, H., Zhang, X. \& Fu, X. A graph convolutional encoder and decoder model for rumor detection. {\em 2020 IEEE 7th International Conference On Data Science And Advanced Analytics (DSAA)}. pp. 300-306 (2020)
\bibitem{ying2018hierarchical}Ying, Z., You, J., Morris, C., Ren, X., Hamilton, W. \& Leskovec, J. Hierarchical graph representation learning with differentiable pooling. {\em Advances In Neural Information Processing Systems}. \textbf{31} (2018)
\bibitem{gao2021topology}Gao, H., Liu, Y. \& Ji, S. Topology-aware graph pooling networks. {\em IEEE Transactions On Pattern Analysis And Machine Intelligence}. \textbf{43}, 4512-4518 (2021)
\bibitem{min2022transformer}Min, E., Chen, R., Bian, Y., Xu, T., Zhao, K., Huang, W., Zhao, P., Huang, J., Ananiadou, S. \& Rong, Y. Transformer for Graphs: An Overview from Architecture Perspective. {\em ArXiv Preprint ArXiv:2202.08455}. (2022)
\bibitem{mikolov2013efficient}Mikolov, T., Chen, K., Corrado, G. \& Dean, J. Efficient estimation of word representations in vector space. {\em ArXiv Preprint ArXiv:1301.3781}. (2013)
\bibitem{perozzi2014deepwalk}Perozzi, B., Al-Rfou, R. \& Skiena, S. Deepwalk: Online learning of social representations. {\em Proceedings Of The 20th ACM SIGKDD International Conference On Knowledge Discovery And Data Mining}. pp. 701-710 (2014)
\bibitem{grover2016node2vec}Grover, A. \& Leskovec, J. node2vec: Scalable feature learning for networks. {\em Proceedings Of The 22nd ACM SIGKDD International Conference On Knowledge Discovery And Data Mining}. pp. 855-864 (2016)
\bibitem{tang2015line}Tang, J., Qu, M., Wang, M., Zhang, M., Yan, J. \& Mei, Q. Line: Large-scale information network embedding. {\em Proceedings Of The 24th International Conference On World Wide Web}. pp. 1067-1077 (2015)
\bibitem{wang2016structural}Wang, D., Cui, P. \& Zhu, W. Structural deep network embedding. {\em Proceedings Of The 22nd ACM SIGKDD International Conference On Knowledge Discovery And Data Mining}. pp. 1225-1234 (2016)
\bibitem{dong2017metapath2vec}Dong, Y., Chawla, N. \& Swami, A. metapath2vec: Scalable representation learning for heterogeneous networks. {\em Proceedings Of The 23rd ACM SIGKDD International Conference On Knowledge Discovery And Data Mining}. pp. 135-144 (2017)
\bibitem{chami2020machine}Chami, I., Abu-El-Haija, S., Perozzi, B., Ré, C. \& Murphy, K. Machine learning on graphs: A model and comprehensive taxonomy. {\em ArXiv Preprint ArXiv:2005.03675}. pp. 1 (2020)
\bibitem{ahmed2013distributed}Ahmed, A., Shervashidze, N., Narayanamurthy, S., Josifovski, V. \& Smola, A. Distributed large-scale natural graph factorization. {\em Proceedings Of The 22nd International Conference On World Wide Web}. pp. 37-48 (2013)
\bibitem{cao2015grarep}Cao, S., Lu, W. \& Xu, Q. Grarep: Learning graph representations with global structural information. {\em Proceedings Of The 24th ACM International On Conference On Information And Knowledge Management}. pp. 891-900 (2015)
\bibitem{ou2016asymmetric}Ou, M., Cui, P., Pei, J., Zhang, Z. \& Zhu, W. Asymmetric transitivity preserving graph embedding. {\em Proceedings Of The 22nd ACM SIGKDD International Conference On Knowledge Discovery And Data Mining}. pp. 1105-1114 (2016)
\bibitem{ying2021transformers}Ying, C., Cai, T., Luo, S., Zheng, S., Ke, G., He, D., Shen, Y. \& Liu, T. Do transformers really perform badly for graph representation?. {\em Advances In Neural Information Processing Systems}. \textbf{34} pp. 28877-28888 (2021)
\bibitem{rong2020self}Rong, Y., Bian, Y., Xu, T., Xie, W., Wei, Y., Huang, W. \& Huang, J. Self-supervised graph transformer on large-scale molecular data. {\em Advances In Neural Information Processing Systems}. \textbf{33} pp. 12559-12571 (2020)
\bibitem{zhang2020graph}Zhang, J., Zhang, H., Xia, C. \& Sun, L. Graph-bert: Only attention is needed for learning graph representations. {\em ArXiv Preprint ArXiv:2001.05140}. (2020)
\bibitem{zhao2021gophormer}Zhao, J., Li, C., Wen, Q., Wang, Y., Liu, Y., Sun, H., Xie, X. \& Ye, Y. Gophormer: Ego-Graph Transformer for Node Classification. {\em ArXiv Preprint ArXiv:2110.13094}. (2021)
\bibitem{tholke2022torchmd}Thölke, P. \& De Fabritiis, G. TorchMD-NET: Equivariant Transformers for Neural Network based Molecular Potentials. {\em ArXiv Preprint ArXiv:2202.02541}. (2022)
\bibitem{wang2022causalgnn}Wang, L., Adiga, A., Chen, J., Sadilek, A., Venkatramanan, S. \& Marathe, M. CausalGNN: Causal-based Graph Neural Networks for Spatio-Temporal Epidemic Forecasting.  (2022)
\bibitem{wang2020fake}Wang, Y., Qian, S., Hu, J., Fang, Q. \& Xu, C. Fake news detection via knowledge-driven multimodal graph convolutional networks. {\em Proceedings Of The 2020 International Conference On Multimedia Retrieval}. pp. 540-547 (2020)
\bibitem{deng2020cola}Deng, S., Wang, S., Rangwala, H., Wang, L. \& Ning, Y. Cola-GNN: Cross-location attention based graph neural networks for long-term ILI prediction. {\em Proceedings Of The 29th ACM International Conference On Information \& Knowledge Management}. pp. 245-254 (2020)
\bibitem{xia2022spatial}Xia, L., Huang, C., Xu, Y., Dai, P., Bo, L., Zhang, X. \& Chen, T. Spatial-Temporal Sequential Hypergraph Network for Crime Prediction. {\em ArXiv Preprint ArXiv:2201.02435}. (2022)
\bibitem{chaolong2018spatio}Chaolong, L., Zhen, C., Wenming, Z., Chunyan, X. \& Jian, Y. Spatio-temporal graph convolution for skeleton based action recognition. {\em Proceedings Of The Irty-Second AAAI Conference On Artificial Intelligence}. (2018)
\bibitem{sawhney2021stock}Sawhney, R., Agarwal, S., Wadhwa, A., Derr, T. \& Shah, R. Stock selection via spatiotemporal hypergraph attention network: A learning to rank approach. {\em Proceedings Of The AAAI Conference On Artificial Intelligence}. \textbf{35}, 497-504 (2021)
\bibitem{wu2020connecting}Wu, Z., Pan, S., Long, G., Jiang, J., Chang, X. \& Zhang, C. Connecting the dots: Multivariate time series forecasting with graph neural networks. {\em Proceedings Of The 26th ACM SIGKDD International Conference On Knowledge Discovery \& Data Mining}. pp. 753-763 (2020)
\bibitem{jain2016structural}Jain, A., Zamir, A., Savarese, S. \& Saxena, A. Structural-rnn: Deep learning on spatio-temporal graphs. {\em Proceedings Of The Ieee Conference On Computer Vision And Pattern Recognition}. pp. 5308-5317 (2016)
\bibitem{ding2021simple}Ding, Z., Ma, Y., He, B. \& Tresp, V. A simple but powerful graph encoder for temporal knowledge graph completion. {\em ArXiv Preprint ArXiv:2112.07791}. (2021)
\bibitem{sun2022ddgcn}Sun, M., Zhang, X., Zheng, J. \& Ma, G. DDGCN: Dual Dynamic Graph Convolutional Networks for Rumor Detection on Social Media.  (2022)
\bibitem{kapoor2020examining}Kapoor, A., Ben, X., Liu, L., Perozzi, B., Barnes, M., Blais, M. \& O'Banion, S. Examining covid-19 forecasting using spatio-temporal graph neural networks. {\em ArXiv Preprint ArXiv:2007.03113}. (2020)
\bibitem{wang2020generic}Wang, Y., Li, P., Bai, C., Subrahmanian, V. \& Leskovec, J. Generic representation learning for dynamic social interaction. {\em Proc. 26th ACM SIGKDD Int. Conf. Knowl. Discovery Data Mining Workshop}. (2020)
\bibitem{seo2018structured}Seo, Y., Defferrard, M., Vandergheynst, P. \& Bresson, X. Structured sequence modeling with graph convolutional recurrent networks. {\em International Conference On Neural Information Processing}. pp. 362-373 (2018)
\bibitem{yu2017spatio}Yu, B., Yin, H. \& Zhu, Z. Spatio-temporal graph convolutional networks: A deep learning framework for traffic forecasting. {\em ArXiv Preprint ArXiv:1709.04875}. (2017)
\bibitem{kim2021learning}Kim, B., Ye, J. \& Kim, J. Learning dynamic graph representation of brain connectome with spatio-temporal attention. {\em Advances In Neural Information Processing Systems}. \textbf{34} pp. 4314-4327 (2021)
\bibitem{li2020dynamic}Li, X., Zhang, M., Wu, S., Liu, Z., Wang, L. \& Philip, S. Dynamic graph collaborative filtering. {\em 2020 IEEE International Conference On Data Mining (ICDM)}. pp. 322-331 (2020)
\bibitem{li2017diffusion}Li, Y., Yu, R., Shahabi, C. \& Liu, Y. Diffusion convolutional recurrent neural network: Data-driven traffic forecasting. {\em ArXiv Preprint ArXiv:1707.01926}. (2017)
\bibitem{ruiz2020gated}Ruiz, L., Gama, F. \& Ribeiro, A. Gated graph recurrent neural networks. {\em IEEE Transactions On Signal Processing}. \textbf{68} pp. 6303-6318 (2020)
\bibitem{tang2021dynamic}Tang, H., Zhao, G., Bu, X. \& Qian, X. Dynamic evolution of multi-graph based collaborative filtering for recommendation systems. {\em Knowledge-Based Systems}. \textbf{228} pp. 107251 (2021)
\bibitem{choi2021dynamic}Choi, J., Ko, T., Choi, Y., Byun, H. \& Kim, C. Dynamic graph convolutional networks with attention mechanism for rumor detection on social media. {\em Plos One}. \textbf{16}, e0256039 (2021)
\bibitem{bian2020rumor}Bian, T., Xiao, X., Xu, T., Zhao, P., Huang, W., Rong, Y. \& Huang, J. Rumor detection on social media with bi-directional graph convolutional networks. {\em Proceedings Of The AAAI Conference On Artificial Intelligence}. \textbf{34} pp. 549-556 (2020)
\bibitem{chen2018gc}Chen, J., Wang, X. \& Xu, X. Gc-lstm: Graph convolution embedded lstm for dynamic link prediction. {\em ArXiv Preprint ArXiv:1812.04206}. (2018)
\bibitem{yan2018spatial}Yan, S., Xiong, Y. \& Lin, D. Spatial temporal graph convolutional networks for skeleton-based action recognition. {\em Thirty-second AAAI Conference On Artificial Intelligence}. (2018)
\bibitem{chen2019lstm}Chen, J., Zhang, J., Xu, X., Fu, C., Zhang, D., Zhang, Q. \& Xuan, Q. E-lstm-d: A deep learning framework for dynamic network link prediction. {\em IEEE Transactions On Systems, Man, And Cybernetics: Systems}. \textbf{51}, 3699-3712 (2019)
\bibitem{fan2022gnn}Fan, J., Bai, J., Li, Z., Ortiz-Bobea, A. \& Gomes, C. A GNN-RNN approach for harnessing geospatial and temporal information: application to crop yield prediction. {\em Proceedings Of The AAAI Conference On Artificial Intelligence}. \textbf{36} pp. 11873-11881 (2022)
\bibitem{kumar2019predicting}Kumar, S., Zhang, X. \& Leskovec, J. Predicting dynamic embedding trajectory in temporal interaction networks. {\em Proceedings Of The 25th ACM SIGKDD International Conference On Knowledge Discovery \& Data Mining}. pp. 1269-1278 (2019)
\bibitem{manessi2020dynamic}Manessi, F., Rozza, A. \& Manzo, M. Dynamic graph convolutional networks. {\em Pattern Recognition}. \textbf{97} pp. 107000 (2020)
\bibitem{wu2019graph}Wu, Z., Pan, S., Long, G., Jiang, J. \& Zhang, C. Graph wavenet for deep spatial-temporal graph modeling. {\em ArXiv Preprint ArXiv:1906.00121}. (2019)
\bibitem{jia2020graphsleepnet}Jia, Z., Lin, Y., Wang, J., Zhou, R., Ning, X., He, Y. \& Zhao, Y. GraphSleepNet: Adaptive Spatial-Temporal Graph Convolutional Networks for Sleep Stage Classification.. {\em IJCAI}. pp. 1324-1330 (2020)
\bibitem{singer2019node}Singer, U., Guy, I. \& Radinsky, K. Node embedding over temporal graphs. {\em ArXiv Preprint ArXiv:1903.08889}. (2019)
\bibitem{zhou2019dynamic}Zhou, Y., Liu, W., Pei, Y., Wang, L., Zha, D. \& Fu, T. Dynamic network embedding by semantic evolution. {\em 2019 International Joint Conference On Neural Networks (IJCNN)}. pp. 1-8 (2019)
\bibitem{nicolicioiu2019recurrent}Nicolicioiu, A., Duta, I. \& Leordeanu, M. Recurrent space-time graph neural networks. {\em Advances In Neural Information Processing Systems}. \textbf{32} (2019)
\bibitem{duta2020dynamic}Duta, I., Nicolicioiu, A. \& Leordeanu, M. Dynamic Regions Graph Neural Networks for Spatio-Temporal Reasoning. {\em ORLR Workshop, Neural Information Processing Systems (NeurIPS)}. (2020)
\bibitem{li2019predicting}Li, J., Han, Z., Cheng, H., Su, J., Wang, P., Zhang, J. \& Pan, L. Predicting path failure in time-evolving graphs. {\em Proceedings Of The 25th ACM SIGKDD International Conference On Knowledge Discovery \& Data Mining}. pp. 1279-1289 (2019)
\bibitem{schlichtkrull2018modeling}Schlichtkrull, M., Kipf, T., Bloem, P., Berg, R., Titov, I. \& Welling, M. Modeling relational data with graph convolutional networks. {\em European Semantic Web Conference}. pp. 593-607 (2018)
\bibitem{behrouz2022anomaly}Behrouz, A. \& Seltzer, M. Anomaly Detection in Multiplex Dynamic Networks: from Blockchain Security to Brain Disease Prediction. {\em ArXiv Preprint ArXiv:2211.08378}. (2022)
\bibitem{pareja2020evolvegcn}Pareja, A., Domeniconi, G., Chen, J., Ma, T., Suzumura, T., Kanezashi, H., Kaler, T., Schardl, T. \& Leiserson, C. Evolvegcn: Evolving graph convolutional networks for dynamic graphs. {\em Proceedings Of The AAAI Conference On Artificial Intelligence}. \textbf{34} pp. 5363-5370 (2020)
\bibitem{goyal2018dyngem}Goyal, P., Kamra, N., He, X. \& Liu, Y. Dyngem: Deep embedding method for dynamic graphs. {\em ArXiv Preprint ArXiv:1805.11273}. (2018)
\bibitem{hajiramezanali2019variational}Hajiramezanali, E., Hasanzadeh, A., Narayanan, K., Duffield, N., Zhou, M. \& Qian, X. Variational graph recurrent neural networks. {\em Advances In Neural Information Processing Systems}. \textbf{32} (2019)
\bibitem{trivedi2019dyrep}Trivedi, R., Farajtabar, M., Biswal, P. \& Zha, H. Dyrep: Learning representations over dynamic graphs. {\em International Conference On Learning Representations}. (2019)
\bibitem{song2021temporally}Song, C., Shu, K. \& Wu, B. Temporally evolving graph neural network for fake news detection. {\em Information Processing \& Management}. \textbf{58}, 102712 (2021)
\bibitem{rossi2020temporal}Rossi, E., Chamberlain, B., Frasca, F., Eynard, D., Monti, F. \& Bronstein, M. Temporal graph networks for deep learning on dynamic graphs. {\em ArXiv Preprint ArXiv:2006.10637}. (2020)
\bibitem{xu2020inductive}Xu, D., Ruan, C., Korpeoglu, E., Kumar, S. \& Achan, K. Inductive representation learning on temporal graphs. {\em ArXiv Preprint ArXiv:2002.07962}. (2020)
\bibitem{qu2020continuous}Qu, L., Zhu, H., Duan, Q. \& Shi, Y. Continuous-time link prediction via temporal dependent graph neural network. {\em Proceedings Of The Web Conference 2020}. pp. 3026-3032 (2020)
\bibitem{wang2021tcl}Wang, L., Chang, X., Li, S., Chu, Y., Li, H., Zhang, W., He, X., Song, L., Zhou, J. \& Yang, H. Tcl: Transformer-based dynamic graph modelling via contrastive learning. {\em ArXiv Preprint ArXiv:2105.07944}. (2021)
\bibitem{chang2020continuous}Chang, X., Liu, X., Wen, J., Li, S., Fang, Y., Song, L. \& Qi, Y. Continuous-time dynamic graph learning via neural interaction processes. {\em Proceedings Of The 29th ACM International Conference On Information \& Knowledge Management}. pp. 145-154 (2020)
\bibitem{wang2021inductive}Wang, Y., Chang, Y., Liu, Y., Leskovec, J. \& Li, P. Inductive representation learning in temporal networks via causal anonymous walks. {\em ArXiv Preprint ArXiv:2101.05974}. (2021)
\bibitem{huang2021temporal}Huang, H., Shi, R., Zhou, W., Wang, X., Jin, H. \& Fu, X. Temporal Heterogeneous Information Network Embedding.. {\em IJCAI}. pp. 1470-1476 (2021)
\bibitem{huang2022learning}Huang, C., Wang, L., Cao, X., Ma, W. \& Vosoughi, S. Learning Dynamic Graph Embeddings Using Random Walk With Temporal Backtracking. {\em NeurIPS 2022 Temporal Graph Learning Workshop}.
\bibitem{yu2020tagnn}Yu, F., Zhu, Y., Liu, Q., Wu, S., Wang, L. \& Tan, T. TAGNN: target attentive graph neural networks for session-based recommendation. {\em Proceedings Of The 43rd International ACM SIGIR Conference On Research And Development In Information Retrieval}. pp. 1921-1924 (2020)
\bibitem{wu2019session}Wu, S., Tang, Y., Zhu, Y., Wang, L., Xie, X. \& Tan, T. Session-based recommendation with graph neural networks. {\em Proceedings Of The AAAI Conference On Artificial Intelligence}. \textbf{33} pp. 346-353 (2019)
\bibitem{zhang2022dynamic}Zhang, M., Wu, S., Yu, X., Liu, Q. \& Wang, L. Dynamic graph neural networks for sequential recommendation. {\em IEEE Transactions On Knowledge And Data Engineering}. (2022)
\bibitem{zhou2020heterogeneous}Zhou, F., Xu, X., Li, C., Trajcevski, G., Zhong, T. \& Zhang, K. A heterogeneous dynamical graph neural networks approach to quantify scientific impact. {\em ArXiv Preprint ArXiv:2003.12042}. (2020)
\bibitem{ivanov2018anonymous}Ivanov, S. \& Burnaev, E. Anonymous walk embeddings. {\em International Conference On Machine Learning}. pp. 2186-2195 (2018)
\bibitem{nguyen2018continuous}Nguyen, G., Lee, J., Rossi, R., Ahmed, N., Koh, E. \& Kim, S. Continuous-time dynamic network embeddings. {\em Companion Proceedings Of The The Web Conference 2018}. pp. 969-976 (2018)
\bibitem{lin2020t}Lin, D., Wu, J., Yuan, Q. \& Zheng, Z. T-edge: Temporal weighted multidigraph embedding for ethereum transaction network analysis. {\em Frontiers In Physics}. \textbf{8} pp. 204 (2020)
\bibitem{khoshraftar2019dynamic}Khoshraftar, S., Mahdavi, S., An, A., Hu, Y. \& Liu, J. Dynamic graph embedding via LSTM History tracking. {\em 2019 IEEE International Conference On Data Science And Advanced Analytics (DSAA)}. pp. 119-127 (2019)
\bibitem{sato2019dyane}Sato, K., Oka, M., Barrat, A. \& Cattuto, C. Dyane: dynamics-aware node embedding for temporal networks. {\em ArXiv Preprint ArXiv:1909.05976}. (2019)
\bibitem{zhou2020graph}Zhou, J., Cui, G., Hu, S., Zhang, Z., Yang, C., Liu, Z., Wang, L., Li, C. \& Sun, M. Graph neural networks: A review of methods and applications. {\em AI Open}. \textbf{1} pp. 57-81 (2020)
\bibitem{wang2021adaptive}Wang, Y., Cai, Y., Liang, Y., Ding, H., Wang, C., Bhatia, S. \& Hooi, B. Adaptive data augmentation on temporal graphs. {\em Advances In Neural Information Processing Systems}. \textbf{34} pp. 1440-1452 (2021)
\bibitem{tiukhova2022influencer}Tiukhova, E., Penaloza, E., Óskarsdóttir, M., Garcia, H., Bahnsen, A., Baesens, B., Snoeck, M. \& Bravo, C. Influencer Detection with Dynamic Graph Neural Networks. {\em ArXiv Preprint ArXiv:2211.09664}. (2022)
\bibitem{kong2020flag}Kong, K., Li, G., Ding, M., Wu, Z., Zhu, C., Ghanem, B., Taylor, G. \& Goldstein, T. Flag: Adversarial data augmentation for graph neural networks. {\em ArXiv Preprint ArXiv:2010.09891}. (2020)
\bibitem{zhao2021data}Zhao, T., Liu, Y., Neves, L., Woodford, O., Jiang, M. \& Shah, N. Data augmentation for graph neural networks. {\em Proceedings Of The AAAI Conference On Artificial Intelligence}. \textbf{35} pp. 11015-11023 (2021)
\bibitem{srivastava2014dropout}Srivastava, N., Hinton, G., Krizhevsky, A., Sutskever, I. \& Salakhutdinov, R. Dropout: a simple way to prevent neural networks from overfitting. {\em The Journal Of Machine Learning Research}. \textbf{15}, 1929-1958 (2014)
\bibitem{he2015deep}He, K., Zhang, X., Ren, S. \& Sun, J. Deep residual learning for image recognition. CoRR abs/1512.03385 (2015).  (2015)
\bibitem{chen2020measuring}Chen, D., Lin, Y., Li, W., Li, P., Zhou, J. \& Sun, X. Measuring and relieving the over-smoothing problem for graph neural networks from the topological view. {\em Proceedings Of The AAAI Conference On Artificial Intelligence}. \textbf{34}, 3438-3445 (2020)
\bibitem{zhao2019pairnorm}Zhao, L. \& Akoglu, L. Pairnorm: Tackling oversmoothing in gnns. {\em ArXiv Preprint ArXiv:1909.12223}. (2019)
\bibitem{cai2020note}Cai, C. \& Wang, Y. A note on over-smoothing for graph neural networks. {\em ArXiv Preprint ArXiv:2006.13318}. (2020)
\bibitem{alon2020bottleneck}Alon, U. \& Yahav, E. On the bottleneck of graph neural networks and its practical implications. {\em ArXiv Preprint ArXiv:2006.05205}. (2020)
\bibitem{topping2021understanding}Topping, J., Di Giovanni, F., Chamberlain, B., Dong, X. \& Bronstein, M. Understanding over-squashing and bottlenecks on graphs via curvature. {\em ArXiv Preprint ArXiv:2111.14522}. (2021)
\bibitem{di2023over}Di Giovanni, F., Giusti, L., Barbero, F., Luise, G., Lio, P. \& Bronstein, M. On Over-Squashing in Message Passing Neural Networks: The Impact of Width, Depth, and Topology. {\em ArXiv Preprint ArXiv:2302.02941}. (2023)
\bibitem{snapnets}Leskovec, J. \& Krevl, A. SNAP Datasets: Stanford Large Network Dataset Collection. (http://snap.stanford.edu/data,2014,6)
\bibitem{madan2011sensing}Madan, A., Cebrian, M., Moturu, S., Farrahi, K. \& Others Sensing the" health state" of a community. {\em IEEE Pervasive Computing}. \textbf{11}, 36-45 (2011)
\bibitem{opsahl2009clustering}Opsahl, T. \& Panzarasa, P. Clustering in weighted networks. {\em Social Networks}. \textbf{31}, 155-163 (2009)
\bibitem{jagadish2014big}Jagadish, H., Gehrke, J., Labrinidis, A., Papakonstantinou, Y., Patel, J., Ramakrishnan, R. \& Shahabi, C. Big data and its technical challenges. {\em Communications Of The ACM}. \textbf{57}, 86-94 (2014)
\bibitem{nr}Rossi, R. \& Ahmed, N. The Network Data Repository with Interactive Graph Analytics and Visualization. {\em AAAI}. (2015), https://networkrepository.com
\bibitem{kumar2016edge}Kumar, S., Spezzano, F., Subrahmanian, V. \& Faloutsos, C. Edge weight prediction in weighted signed networks. {\em Data Mining (ICDM), 2016 IEEE 16th International Conference On}. pp. 221-230 (2016)
\bibitem{kumar2018rev2}Kumar, S., Hooi, B., Makhija, D., Kumar, M., Faloutsos, C. \& Subrahmanian, V. Rev2: Fraudulent user prediction in rating platforms. {\em Proceedings Of The Eleventh ACM International Conference On Web Search And Data Mining}. pp. 333-341 (2018)
\bibitem{leskovec2005graphs}Leskovec, J., Kleinberg, J. \& Faloutsos, C. Graphs over time: densification laws, shrinking diameters and possible explanations. {\em Proceedings Of The Eleventh ACM SIGKDD International Conference On Knowledge Discovery In Data Mining}. pp. 177-187 (2005)
\bibitem{yang2012defining}Yang, J. \& Leskovec, J. Defining and Evaluating Network Communities based on Ground-truth. Extended version. (Citeseer,2012)
\bibitem{kumar2021deception}Kumar, S., Bai, C., Subrahmanian, V. \& Leskovec, J. Deception Detection in Group Video Conversations using Dynamic Interaction Networks. {\em ICWSM 2021}. (2021)
\bibitem{bai2019predicting}Bai, C., Kumar, S., Leskovec, J., Metzger, M., Nunamaker, J. \& Subrahmanian, V. Predicting the Visual Focus of Attention in Multi-Person Discussion Videos. {\em IJCAI 2019}. (2019)
\bibitem{stuart2010artificial}Stuart, J. \& Others Artificial Intelligence A Modern Approach Third Edition. (Prentice Hall,2010)
\bibitem{dietterich2002machine}Dietterich, T. Machine learning for sequential data: A review. {\em Joint IAPR International Workshops On Statistical Techniques In Pattern Recognition (SPR) And Structural And Syntactic Pattern Recognition (SSPR)}. pp. 15-30 (2002)
\bibitem{hamilton2020graph}Hamilton, W. Graph representation learning. {\em Synthesis Lectures On Artifical Intelligence And Machine Learning}. \textbf{14}, 1-159 (2020)
\bibitem{bengio2013representation}Bengio, Y., Courville, A. \& Vincent, P. Representation learning: A review and new perspectives. {\em IEEE Transactions On Pattern Analysis And Machine Intelligence}. \textbf{35}, 1798-1828 (2013)
\bibitem{bahmani2012pagerank}Bahmani, B., Kumar, R., Mahdian, M. \& Upfal, E. Pagerank on an evolving graph. {\em Proceedings Of The 18th ACM SIGKDD International Conference On Knowledge Discovery And Data Mining}. pp. 24-32 (2012)
\bibitem{gao2022equivalence}Gao, J. \& Ribeiro, B. On the equivalence between temporal and static equivariant graph representations. {\em International Conference On Machine Learning}. pp. 7052-7076 (2022)
\bibitem{van2020survey}Van Engelen, J. \& Hoos, H. A survey on semi-supervised learning. {\em Machine Learning}. \textbf{109}, 373-440 (2020)
\bibitem{kalofolias2017learning}Kalofolias, V., Loukas, A., Thanou, D. \& Frossard, P. Learning time varying graphs. {\em 2017 IEEE International Conference On Acoustics, Speech And Signal Processing (ICASSP)}. pp. 2826-2830 (2017)
\bibitem{casteigts2012time}Casteigts, A., Flocchini, P., Quattrociocchi, W. \& Santoro, N. Time-varying graphs and dynamic networks. {\em International Journal Of Parallel, Emergent And Distributed Systems}. \textbf{27}, 387-408 (2012)
\bibitem{wang2019time}Wang, Y., Yuan, Y., Ma, Y. \& Wang, G. Time-dependent graphs: Definitions, applications, and algorithms. {\em Data Science And Engineering}. \textbf{4} pp. 352-366 (2019)
\bibitem{huang2020reliable}Huang, H., Savkin, A. \& Huang, C. Reliable path planning for drone delivery using a stochastic time-dependent public transportation network. {\em IEEE Transactions On Intelligent Transportation Systems}. \textbf{22}, 4941-4950 (2020)
\bibitem{li2022enhanced}Li, J., Wang, P., Li, H. \& Shi, K. Enhanced time-expanded graph for space information network modeling. {\em Science China Information Sciences}. \textbf{65}, 192301 (2022)
\bibitem{azimifar2015vehicle}Azimifar, M., Todd, T., Khezrian, A. \& Karakostas, G. Vehicle-to-vehicle forwarding in green roadside infrastructure. {\em IEEE Transactions On Vehicular Technology}. \textbf{65}, 780-795 (2015)
\bibitem{rehman2020exploring}Rehman, A., Ahmad, M. \& Khan, O. Exploring accelerator and parallel graph algorithmic choices for temporal graphs. {\em Proceedings Of The Eleventh International Workshop On Programming Models And Applications For Multicores And Manycores}. pp. 1-10 (2020)
\bibitem{gilmer17:_neural_messag_passin_quant_chemis}Gilmer, J., Schoenholz, S., Riley, P., Vinyal, O. \& Dahl, G. Neural Message Passing from Quantum Chemistry. {\em Proceedings Of The International Conference On Machine Learning}. (2017)
\bibitem{defferrard16:_convol_neural_networ_graph_fast}Defferrard, M., Bresson, X. \& Vandergheynst, P. Convolutional Neural Networks on Graphs with Fast Localized Spectral Filtering. {\em Advances In Neural Information Processing Systems}. pp. 3844-3852 (2016)
\bibitem{velivckovic2017graph}Veličković, P., Cucurull, G., Casanova, A., Romero, A., Lio, P. \& Bengio, Y. Graph attention networks. {\em International Conference On Learning Representations (ICLR)}. (2018)
\bibitem{bresson2017residual}Residual gated graph convnets. X Bresson, T Laurent, arXiv preprint arXiv:1711.07553, 2017
\bibitem{xu2018how}Xu, K., Hu, W., Leskovec, J. \& Jegelka, S. How Powerful are Graph Neural Networks?. {\em International Conference On Learning Representations}. (2019)
\bibitem{wu2020comprehensive}Wu, Z., Pan, S., Chen, F., Long, G., Zhang, C. \& Philip, S. A comprehensive survey on graph neural networks. {\em IEEE Transactions On Neural Networks And Learning Systems}. \textbf{32}, 4-24 (2020)
\bibitem{morris2019wl}Morris, C., Ritzert, M., Fey, M., Hamilton, W., Lenssen, J., Rattan, G. \& Grohe, M. Weisfeiler and Lehman Go Neural: Higher-Order Graph Neural Networks. {\em Proceedings Of The Thirty-Third AAAI Conference On Artificial Intelligence And Thirty-First Innovative Applications Of Artificial Intelligence Conference And Ninth AAAI Symposium On Educational Advances In Artificial Intelligence}. (2019), https://doi.org/10.1609/aaai.v33i01.33014602
\bibitem{breakingthelimits}Balcilar, M., Héroux, P., Gauzere, B., Vasseur, P., Adam, S. \& Honeine, P. Breaking the limits of message passing graph neural networks. {\em International Conference On Machine Learning}. pp. 599-608 (2021)
\bibitem{maron2019provably}Maron, H., Ben-Hamu, H., Serviansky, H. \& Lipman, Y. Provably powerful graph networks. {\em Advances In Neural Information Processing Systems}. \textbf{32} (2019)
\bibitem{dwivedi2020benchmarking}Dwivedi, V., Joshi, C., Laurent, T., Bengio, Y. \& Bresson, X. Benchmarking graph neural networks. {\em ArXiv Preprint ArXiv:2003.00982}. (2020)
\bibitem{Sutskever14}Sutskever, I., Vinyals, O. \& Le, Q. Sequence to Sequence Learning with Neural Networks. {\em Advances In Neural Information Processing Systems}. pp. 3104-3112 (2014)
\bibitem{Niu21}Niu, Z., Zhong, G. \& Yu, H. A review on the attention mechanism of deep learning. {\em Neurocomputing}. \textbf{452} pp. 48-62 (2021)
\bibitem{self18}Shaw, P., Uszkoreit, J. \& Vaswani, A. Self-Attention with Relative Position Representations. {\em Proceedings Of The 2018 Conference Of The North American Chapter Of The Association For Computational Linguistics: Human Language Technologies, Volume 2 (Short Papers)}. pp. 464-468 (2018)
\bibitem{blstm2013}Graves, A., Mohamed, A. \& Hinton, G. Speech recognition with deep recurrent neural networks. {\em IEEE International Conference On Acoustics, Speech And Signal Processing, ICASSP}. pp. 6645-6649 (2013)
\bibitem{dilated}Yu, F. \& Koltun, V. Multi-Scale Context Aggregation by Dilated Convolutions. {\em 4th International Conference On Learning Representations ICLR }. (2016)


\end{thebibliography}


\end{document}